\def\year{2022}\relax
\documentclass[letterpaper]{article} 

\usepackage{cancel,soul}
\usepackage{multirow}
\newcommand\Tstrut{\rule{0pt}{2.2ex}}         
\newcommand\Bstrut{\rule[-0.7ex]{0pt}{0pt}}   
\usepackage{amsmath}
\usepackage{xcolor}
\usepackage{bm}
\usepackage{microtype}
\usepackage{subfigure}
\usepackage{booktabs} 
\usepackage{amsthm}

\DeclareMathOperator*{\argmax}{argmax} 

\usepackage{clom22}  
\usepackage{times}  
\usepackage{helvet}  
\usepackage{courier}  
\usepackage[hyphens]{url}  
\usepackage{graphicx} 
\usepackage{natbib}  
\usepackage{caption} 
\DeclareCaptionStyle{ruled}{labelfont=normalfont,labelsep=colon,strut=off} 
\usepackage{algorithm}
\usepackage{algorithmic}

\usepackage{newfloat}
\usepackage{listings}
\floatstyle{ruled}
\newfloat{listing}{tb}{lst}{}
\floatname{listing}{Listing}
\title{Continual Learning Based on OOD Detection and Task Masking}
\author {
    Gyuhak Kim,\textsuperscript{\rm 1}
    Sepideh Esmaeilpour,\textsuperscript{\rm 1}
    Changnan Xiao,\textsuperscript{\rm 2}
    Bing Liu\textsuperscript{\rm 1}
}
\affiliations {
    \textsuperscript{\rm 1} University of Illinois at Chicago\\
    \textsuperscript{\rm 2} ByteDance\\
    \{gkim87, sesmae2, liub\}@uic.edu, xiaochangnan@bytedance.com
    
}

\begin{document}

\maketitle

\begin{abstract}
Existing continual learning techniques focus on either \textit{task incremental learning} (TIL) or \textit{class incremental learning} (CIL) {problem}, but not both. CIL and TIL differ mainly in that the task-id is provided for each test sample during testing for TIL, but not provided for CIL. 
{Continual learning methods intended for one problem have limitations on the other problem.}
This paper proposes a novel unified approach based on out-of-distribution (OOD) detection and task masking, called CLOM, to solve both problems.
The key novelty is that each task is trained as an OOD detection model rather than a traditional supervised learning model, and a task mask is trained to protect each task to prevent forgetting. 
Our evaluation shows that CLOM outperforms existing state-of-the-art baselines by large margins. The average TIL/CIL accuracy of CLOM over six experiments is 87.6/67.9\% while that of the best baselines is only 82.4/55.0\%. The code of our system is available at \url{https://github.com/k-gyuhak/CLOM}.
\end{abstract}

\section{Introduction}
\label{sec.intro}
{Continual learning} (CL) learns a sequence of tasks $<$$1, 2, ..., k, ...$$>$ incrementally. Each task $k$ has its dataset $\mathcal{D}_k=\{(\boldsymbol{x}_k^i, y_k^i)_{i=1}^{n_k}\}$, where $\boldsymbol{x}_k^i \in \boldsymbol{X}$ is a data sample in task $k$ and $y_k^i \in \boldsymbol{Y}_k$ is the class label of $\boldsymbol{x}_k^i$ and $\boldsymbol{Y}_k$ is the set of classes of task $k$. The key challenge of CL is \textit{catastrophic forgetting} (CF)~\cite{McCloskey1989}, which refers to the situation where the learning of a new task may significantly change the network weights learned for old tasks, degrading the model accuracy for old tasks. Researchers have mainly worked on two CL problems: \textit{class incremental/continual learning} (CIL) and \textit{task incremental/continual learning} (TIL)~\citep{Dhar2019CVPR,van2019three}.
The main difference between CIL and TIL is that in TIL, the task-id $k$ is provided for each test sample $\boldsymbol{x}$ during testing so that only the model for task $k$ is used to classify $\boldsymbol{x}$, while in CIL, the task-id $k$ for each test sample is not provided.

Existing CL techniques focus on either CIL or TIL~\citep{Parisi2019continual}. In general, CIL methods are designed to function without given task-id to perform class prediction over all classes in the tasks learned so far, and thus tend to forget the previous task performance due to plasticity for the new task. TIL methods are designed to function with a task-id for class prediction within the task. They are more stable to retain previous within-task knowledge, but incompetent if task-id is unknown (see Sec.~\ref{experiment}).

This paper proposes a novel and unified method called CLOM (\textit{C}ontinual \textit{L}earning based on \textit{O}OD detection and Task \textit{M}asking) to solve both problems by overcoming the limitations of CIL and TIL methods. 
CLOM has two key mechanisms: (1) a task mask mechanism for protecting each task model to overcome CF, and (2) a learning method for building a model for each task based on out-of-distribution (OOD) detection. 
The task mask mechanism is inspired by \textit{hard attention} in the TIL system HAT~\cite{Serra2018overcoming}.
The OOD detection based learning method for building each task model is quite different from the classic supervised learning used in existing TIL systems. It is, in fact, the key novelty and enabler for CLOM to work effectively for CIL. 

OOD detection is stated as follows~\cite{bulusu2020anomalous}: Given a training set of $n$ classes, called the \textit{in-distribution} (IND) training data, we want to build a model that can assign correct classes to IND test data and reject or detect OOD test data that do not belong to any of the $n$ IND training classes. The OOD rejection capability makes a TIL model effective for CIL problem because during testing, if a test sample does not belong to any of the classes of a task, it will be rejected by the model of the task. Thus, only the task that the test sample belongs to will accept it and classify it to one of its classes. No task-id is needed. The OOD detection algorithm in CLOM is inspired by several recent advances in self-supervised learning, data augmentation~\cite{he2020momentum}, contrastive learning~\cite{oord2018representation,chen2020simple}, and their applications to OOD detection~\cite{golan2018deep,hendrycks2019using,tack2020csi}.

{The main contributions of this work are as follows:
\begin{itemize}
    \item It proposes a novel CL method CLOM, which is essentially a TIL system in training, but solves both TIL and CIL problems in testing. Existing methods mainly solve either TIL or CIL problem, but are weak on the other problem.
    \item CLOM uses task masks to protect the model for each task to prevent CF and OOD detection to build each task model, which to our knowledge has not been done before. More importantly, since the task masks can prevent CF,
    {continual learning performance gets better as OOD models get better too.}
    \item CLOM needn't to save any replay data. If some past data is saved for output calibration, it performs even better. 
\end{itemize}} 
Experimental results show that CLOM improves state-of-the-art baselines by large margins.~The average TIL/CIL accuracy of CLOM over six different experiments is 87.6/67.9\% while that of the best baseline is only 82.4/55.0\%.

\section{Related Work}

Many approaches have been proposed to deal with CF in CL. Using regularization~\cite{kirkpatrick2017overcoming} and knowledge distillation~\cite{Li2016LwF} to minimize the change to previous models are two popular approaches~\cite{Jung2016less,Camoriano2017incremental,Fernando2017pathnet,Rannen2017encoder,Seff2017continual,zenke2017continual,Kemker2018fearnet,ritter2018online,schwarz2018progress,xu2018reinforced,castro2018end,Dhar2019CVPR,hu2019overcoming,lee2019overcoming,Liu2020}. 
Memorizing some old examples and using them to adjust the old models in learning a new task is another popular approach (called \textit{replay})~\cite{Rusu2016,Lopez2017gradient,Rebuffi2017,Chaudhry2019ICLR,Cyprien2019episodic, hou2019learning,wu2019large,rolnick2019neurIPS, NEURIPS2020_b704ea2c_derpp,zhao2020maintaining,rajasegaran2020adaptive,Liu2020AANets}. 
Several systems learn to generate pseudo training data of old tasks and use them to jointly train the new task, called \textit{pseudo-replay}~\cite{Gepperth2016bio,Kamra2017deep,Shin2017continual,wu2018memory,Seff2017continual,wu2018memory,Kemker2018fearnet,hu2019overcoming,hayes2019remind,Rostami2019ijcai,ostapenko2019learning}. CLOM differs from these approaches {as it does not replay any old task data to prevent forgetting and can function with/without saving some old data.} {\textit{Parameter isolation} is yet another popular approach, which makes different subsets (which may overlap) of the model parameters dedicated to different tasks using masks~\cite{Fernando2017pathnet,Serra2018overcoming,ke2020continual} or finding a sub-network for each task by pruning \cite{Mallya2017packnet, supsup2020,NEURIPS2019_3b220b43compact}. CLOM uses parameter isolation, but it differs from these approaches as it combines the idea of parameter isolation and OOD detection, which can solve both TIL and CIL problems effectively.} 

There are also many other approaches, e.g., a network of experts~\cite{Aljundi2016expert} and generalized CL~\cite{mi2020generalized}, etc. {PASS~\cite{Zhu_2021_CVPR_pass} uses data rotation and regularizes them.  
Co$^2$L~\cite{Cha_2021_ICCV_co2l} is a replay method that uses contrastive loss on old samples. CLOM also uses rotation and constrative loss, but its CF handling is based on masks. None of the existing methods uses OOD detection.}

CLOM is a TIL method that can also solve CIL problems. Many TIL systems exist~\cite{Fernando2017pathnet}, e.g., GEM~\cite{Lopez2017gradient}, A-GEM~\cite{Chaudhry2019ICLR}, UCL~\cite{ahn2019neurIPS}, ADP~\cite{Yoon2020Scalable}, CCLL~\cite{singh2020calibrating}, Orthog-Subspace~\cite{chaudhry2020continual}, HyperNet~\cite{von2019continual}, PackNet~\cite{Mallya2017packnet}, CPG~\cite{NEURIPS2019_3b220b43compact}, SupSup~\cite{supsup2020}, HAT~\cite{Serra2018overcoming}, and CAT~\cite{ke2020continual}. 
GEM is a replay based method and UCL is a regularization based method. A-GEM~\citep{Chaudhry2019ICLR} improves GEM's efficiency. {ADP decomposes parameters into shared and adaptive parts to construct an order robust TIL system.} CCLL uses task-adaptive calibration on convolution layers. Orthog-Subspace learns each task in subspaces orthogonal to each other. {HyperNet initializes task-specific parameters conditioned on task-id. PackNet, CPG and SupSup find an isolated sub-network for each task and use it at inference.
HAT and CAT protect previous tasks by masking the important parameters.
Our CLOM also uses this general approach, but its model building for each task is based on OOD detection, which has not been used by existing TIL methods. It also performs well in the CIL setting (see Sec.~\ref{resultcomparison}). } 

Related work on out-of-distribution (OOD) detection (also called open set detection) is also extensive. 
Excellent surveys include~\cite{bulusu2020anomalous,geng2020recent}. 
The model building part of CLOM is inspired by the latest method in~\cite{tack2020csi} based on data augmentation~\cite{he2020momentum} and contrastive learning~\cite{chen2020simple}.

\section{Proposed CLOM Technique}
\label{sec.CLOM}
\begin{figure}
\centering
\includegraphics[width=3in]{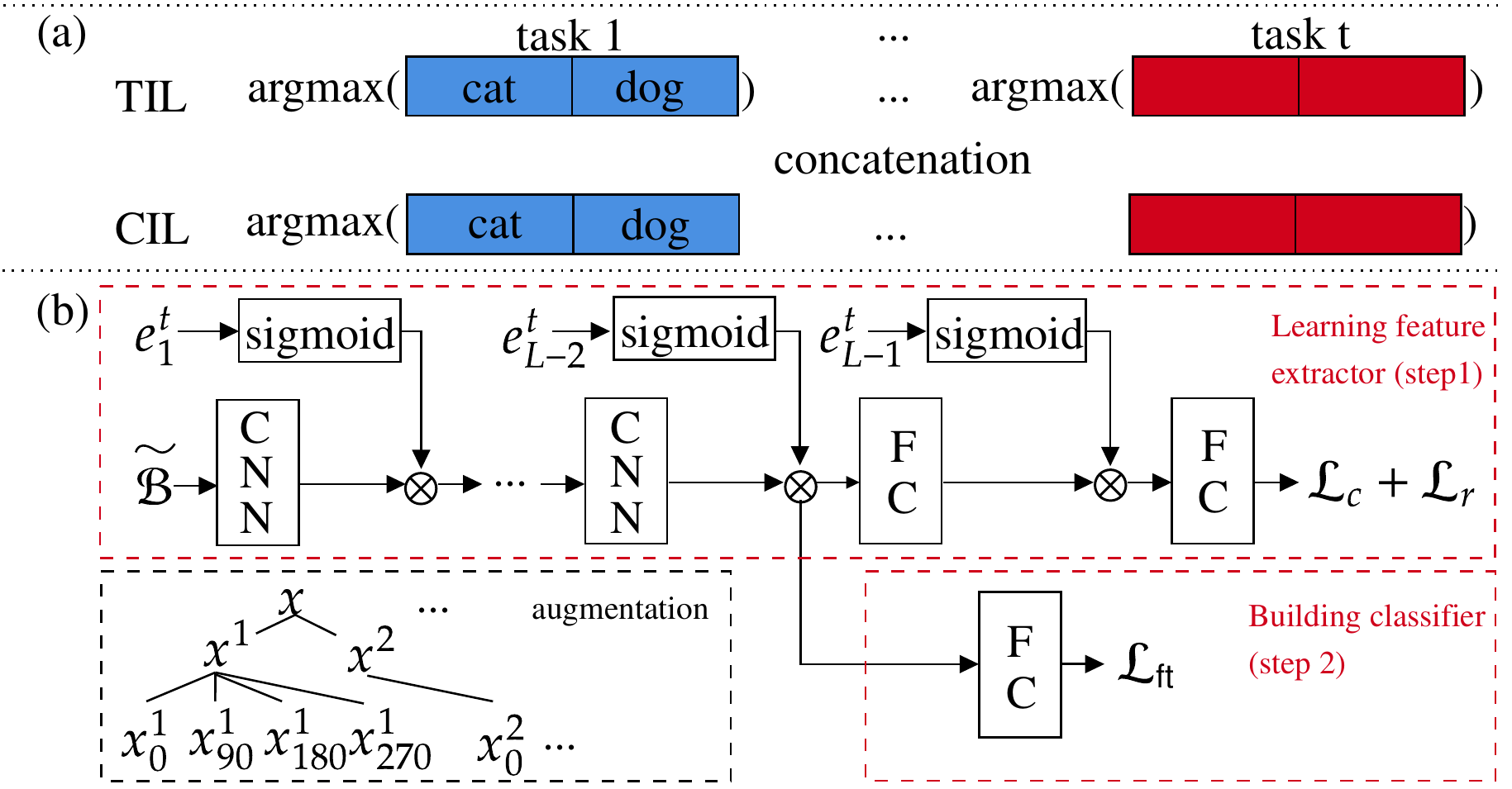}
\caption{(a) Overview of TIL and CIL prediction. For TIL prediction, we consider only the output heads of the given task. For CIL prediction, we obtain all the outputs from task $1$ to current task $t$, and choose the label over the concatenated output.
(b) Training overview for a generic backbone. Each CNN includes a convolution layer with batch normalization, activation, or pooling depending on the configuration. In our experiments, we use AlexNet~\cite{NIPS2012_c399862d} and ResNet-18~\cite{he2016deep}. We feed-forward augmented batch $\tilde{\mathcal{B}}$ consisting of rotated images of different views ($\boldsymbol{x}^{1}_{0}, \boldsymbol{x}^{1}_{90}, \cdots$). We first train feature extractor by using contrastive loss (step 1). At each layer, binary mask is multiplied to the output of each convolution layer to learn important parameters of current task $t$. After training feature extractor, we fine-tune the OOD classifier (step 2).
}
\label{diagram}
\end{figure}

As mentioned earlier, CLOM is a TIL system that solves both TIL and CIL problems.
It takes the parameter isolation approach for TIL. For each task $k$ (task-id), a model is built $f(h(\boldsymbol{x}, k))$ for the task, {where $h$ and $f$ are the feature extractor and the task specific classifier, respectively,}
and $f(h(\boldsymbol{x}, k))$ is the output of the neural network for task $k$. 
{We omit task-id $k$ in $f$ to simplify notation.}
{In learning each task $k$, a task mask is also trained at the same time to protect the learned model of the task. 

In testing for TIL, given the test sample $\boldsymbol{x}$ with task-id $k$ provided, CLOM uses the model for task $k$ to classify $\boldsymbol{x}$,
\begin{align}
    \hat{y} = \argmax 
    f(h(\boldsymbol{x}, k)) 
    \label{TCLeq}
\end{align}
For CIL (no task-id for each test sample $\boldsymbol{x}$), CLOM uses the model of every task and predicts the class $\hat{y}$ using
\begin{align}
    \hat{y} = \argmax 
    \bigoplus_{1\leq k \leq t}  f(h(\boldsymbol{x}, k)),
    \label{CLOMeq}
\end{align}
where
$\bigoplus$ is the concatenation over the output space and
$t$ is the last task that has been learned. Eq.~\ref{CLOMeq} essentially chooses the class with the highest classification score among the classes of all tasks learned so far. This works because the OOD detection model for a task will give very low score for a test sample that does not belong to the task. {An overview of the prediction is illustrated in Fig.~\ref{diagram}(a).}
}

Note that in fact any CIL model can be used as a TIL model if task-id is provided. The conversion is made by selecting the corresponding task classification heads in testing. However, a conversion from a TIL method to a CIL method is not obvious as the system requires task-id in testing.
Some attempts have been made to predict the task-id in order to make a TIL method applicable to CIL problems.
iTAML~\cite{rajasegaran2020itaml} requires the test samples to come in batches and each batch must be from a single task. This may not be practical as test samples usually come one by one. CCG~\cite{abati2020conditional} builds a separate network to predict the task-id, which is also prone to forgetting. Expert Gate~\cite{Aljundi2016expert} constructs a separate autoencoder for each task. 
Our CLOM classifies one test sample at a time and does not need to {construct another network for task-id prediction.}

{In the following subsections, we present (1) how to build an OOD model for a task and use it to make a prediction, and (2) how to learn task masks to protect each model. An overview of the process is illustrated in Fig. \ref{diagram}(b).}

\subsection{Learning Each Task as OOD Detection} \label{unifiedappr}
We borrow the latest OOD ideas based on contrastive learning~\cite{chen2020simple, he2020momentum} and data augmentation due to their excellent performance~\cite{tack2020csi}. 
Since this section focuses on how to learn a single task based on OOD detection, we omit the task-id unless necessary. The OOD training process {is similar to that of contrastive learning}. It consists of two steps: 1) learning {the feature representation by} the composite $g \circ h$, where $h$ is a feature extractor and $g$ is a projection to contrastive representation, and 2) learning a linear classifier $f$ mapping the feature representation of $h$ to the label space. {In the following, we describe the training process: contrastive learning for feature representation learning (1), and OOD classifier building (2). We then explain how to make a prediction based on an ensemble method for both TIL and CIL settings, and how to further improve prediction using some saved past data.}

\subsubsection{Contrastive Loss for Feature Learning.}
{This is step 1 in Fig.~\ref{diagram}(b).} Supervised contrastive learning is used to try to repel data of different classes and align data of the same class more closely to make it easier to classify them. A key operation is data augmentation via transformations. 

{Given a batch of $N$ samples, each sample $\boldsymbol{x}$ is first duplicated and each version then goes through \textit{three initial augmentations}
(also see Data Augmentation in Sec.~\ref{sec:traindetails}) to generate two different views $\boldsymbol{x}^{1}$ and $\boldsymbol{x}^{2}$ (they keep the same class label as $\boldsymbol{x}$).}
Denote the augmented batch by $\mathcal{B}$, which now has $2N$ samples. {In~\cite{hendrycks2019using,tack2020csi}}, it was shown that using image rotations is effective in learning OOD detection models because such rotations can effectively serve as out-of-distribution (OOD) training data.  
For each augmented sample $\boldsymbol{x} \in \mathcal{B}$ with class $y$ of a task, we rotate $\boldsymbol{x}$ by $90^{\circ}, 180^{\circ}, 270^{\circ}$ to create three images, which are assigned \textit{three new classes} $y_1, y_2$, and $y_3$, respectively.
{This results in a larger augmented batch $\tilde{\mathcal{B}}$. Since we generate three new images from each $\boldsymbol{x}$, 
the size of $\tilde{\mathcal{B}}$ is $8N$. For each original class, we now have 4 classes. For a sample $\boldsymbol{x} \in \tilde{\mathcal{B}}$, let $\mathcal{\tilde{B}}(\boldsymbol{x}) = \mathcal{\tilde{B}} \backslash \{ \boldsymbol{x} \}$
and 
let $P(\boldsymbol{x}) \subset \tilde{\mathcal{B}} \backslash \{ \boldsymbol{x} \}$ 
be a set consisting of the data of the same class as $\boldsymbol{x}$ distinct from $\boldsymbol{x}$.
The contrastive representation of a sample $\boldsymbol{x}$ is $\boldsymbol{z}_{x} = g(h(\boldsymbol{x}, t)) / \| g(h(\boldsymbol{x}, t)) \|$, where $t$ is the current task. 
In learning, we minimize the supervised contrastive loss~\cite{khosla2020supervised} of task $t$. 
\begin{align}
    \mathcal{L}_{c}
    &= \frac{1}{8N} \sum_{ \boldsymbol{x} \in \tilde{\mathcal{B}}} \frac{-1}{| P(\boldsymbol{x}) |} \times  \nonumber \\
    & \sum_{\boldsymbol{p} \in P(\boldsymbol{x})} \log{ \frac{ \text{exp}( \boldsymbol{z}_{\boldsymbol{x}} \cdot \boldsymbol{z}_{\boldsymbol{p}} / \tau)}{\sum_{\boldsymbol{x}'  \in \tilde{\mathcal{B}}(\boldsymbol{x}) } \text{exp}( \boldsymbol{z}_{\boldsymbol{x}} \cdot \boldsymbol{z}_{\boldsymbol{x}'} / \tau) } } \label{modsupclr}
\end{align}
where 
$\tau$ is a scalar temperature, $\cdot$ is dot product, and $\times$ is  multiplication. 
The loss is reduced by repelling $\boldsymbol{z}$ of different classes and aligning $\boldsymbol{z}$ of the same class more closely.
$\mathcal{L}_{c}$ basically trains a feature extractor with good 
representations for learning an  OOD classifier.}

\subsubsection{Learning the Classifier.} 
{This is step 2 in Fig.\ref{diagram}(b)}. {Given the feature extractor $h$ trained with the loss in Eq.~\ref{modsupclr}, we {\textit{freeze $h$} and} only \textit{fine-tune} the linear classifier $f$, which is trained to predict the classes of task $t$ \textit{and} the augmented rotation classes.} $f$ maps the feature representation to {the label space in} $\mathcal{R}^{4|\mathcal{C}^{t}|}$, where $4$ is the number of rotation classes including the original data with $0^{\circ}$ rotation and $|\mathcal{C}^{t}|$ is the number of {original} classes in task $t$. We minimize the cross-entropy loss,
\begin{align}
    \mathcal{L}_{\text{ft}} = - \frac{1}{|\tilde{\mathcal{B}} |} \sum_{(\boldsymbol{x}, y) \in \tilde{\mathcal{B}}} 
    \log \tilde{p}(y | \boldsymbol{x}, t) 
    \label{3obj}
\end{align}
where $\text{ft}$ indicates fine-tune,
and 
\begin{align}
    \tilde{p}(y | \boldsymbol{x}, t) = \text{softmax} \left( f(h(\boldsymbol{x}, t))
    \right) \label{probrotation}
\end{align}
where
$f(h(\boldsymbol{x}, t)) \in \mathcal{
R}^{4|\mathcal{C}^{t}|}$. The output $f(h(\boldsymbol{x}, t))$
includes the rotation classes. The linear classifier is trained to predict the original \textit{and} the rotation classes.

\subsubsection{Ensemble Class Prediction.} \label{ensemblesection}
{
We describe how to predict a label $y \in \mathcal{C}^{t}$ (TIL) and $y \in \mathcal{C}$ (CIL) ($\mathcal{C}$ is the set of original classes of all tasks)
We assume all tasks have been learned and their models are protected by masks, which we discuss in the next subsection. 

We discuss the prediction of class label $y$ for a test sample $\boldsymbol{x}$ in the TIL setting first. Note that the network $f\circ h$ in Eq.~\ref{probrotation} returns logits for rotation classes (including the original task classes). Note also for each original class label $j_k \in \mathcal{C}^{k}$ (original classes) of a task $k$, we created three additional rotation classes. For class $j_k$, the classifier $f$ will produce four output values from its four rotation class logits, i.e., $f_{j_k,0}(h(\boldsymbol{x_0}, k))$, $f_{j_k,90}(h(\boldsymbol{x_{90}}, k))$, $f_{j_k,180}(h(\boldsymbol{x_{180}}, k))$, and $f_{j_k,270}(h(\boldsymbol{x_{270}}, k))$, where 0, 90, 180, and 270 represent $0^{\circ}, 90^{\circ}, 180^{\circ}$, and $270^{\circ}$ rotations respectively and $\boldsymbol{x}_0$ is the original $\boldsymbol{x}$. 
We compute an ensemble output $f_{j_k}(h(\boldsymbol{x},k))$ for each class $j_k \in \mathcal{C}^{k}$ of task $k$, 
\begin{align}
    f_{j_k}(h(\boldsymbol{x},k))= \frac{1}{4} \sum_{\text{deg}} f_{j_k,\text{deg}} (h(\boldsymbol{x}_{\text{deg}}, k)) \label{ensemblelogit}.
\end{align}
The final TIL class prediction is made as follows (note, in TIL, task-id $k$ is provided in testing),
\begin{align}
    \hat{y} = \argmax_{j_k \in \mathcal{C}^{k}}  \Big\{ f_{j_k}(h(\boldsymbol{x},k)) \Big\} \label{ensembleclass}
\end{align} 
We use Eq.~\ref{CLOMeq} to make the CIL class prediction, where {the final format} $f(h(\boldsymbol{x},k))$ for task $k$ is the following vector: 
\begin{align}
   f(h(\boldsymbol{x}, k)) = \left[ f_{1}(h(\boldsymbol{x}, k), \cdots , f_{|\mathcal{C}^{k}|}(h(\boldsymbol{x}, k)) \right]
\label{eq:afterensemble}
\end{align}
Our method so far memorizes no training samples and it already outperforms baselines (see Sec.~\ref{resultcomparison}). 
}

\subsubsection{Output Calibration with Memory.}
\label{sec.calibration}
{The proposed method can make incorrect CIL prediction even with a perfect OOD model (rejecting every test sample that does not belong any class of the task). This happens because the task models are trained independently, and the outputs of different tasks may have different magnitudes. We use output calibration to ensure that the outputs are of similar magnitudes to be comparable by using some saved examples in a memory $\mathcal{M}$ with limited budget.}
At each task $k$, we store a fraction of the validation data $\{ (\boldsymbol{x}, y)\}$ into $\mathcal{M}_k$ for output calibration and update the memory $\mathcal{M} \leftarrow update(\mathcal{M}, \mathcal{M}_k)$ {by maintaining an equal number of samples per class}. We will detail the memory budget in Sec.~\ref{ablation}. Basically, we save the same number as the existing replay-based TIL/CIL methods. 

After training the network $f$ for the current task $t$,
we freeze the model and use the saved data in $\mathcal{M}$ to find the scaling and shifting parameters $(\boldsymbol{\sigma}, \boldsymbol{\mu}) \in \mathcal{R}^t \times \mathcal{R}^t$ to calibrate the after-ensemble classification output $f(h(\boldsymbol{x},k))$ (Eq.~\ref{eq:afterensemble}) (i.e., using $\sigma_{k} f(h (\boldsymbol{x}, k)) + \mu_k)$ for each task $k$ by minimizing the cross-entropy loss,
\begin{align}
    \mathcal{L}_{\text{cal}}
    = - \frac{1}{| \mathcal{M} |} \sum_{(\boldsymbol{x}, y) \in \mathcal{M}}
    \log{ p(y| \boldsymbol{x}) }
\end{align}
where $\mathcal{C}$ is the set of all classes of all tasks seen so far, and $p(c|\boldsymbol{x})$ is computed using softmax,
\begin{align}
\text{softmax} \Big[  \sigma_{1} f(h (\boldsymbol{x}, 1)) + \mu_1 ;
   \cdots ; 
    \sigma_{t} f(h (\boldsymbol{x}, t)) + \mu_t \Big]
    \label{eq:scaling}
\end{align}
Clearly, parameters $\sigma_k$ and $\mu_k$ do not change classification within task $k$ (TIL), but calibrates the outputs such that the ensemble outputs from all tasks are in comparable magnitudes.
For CIL inference at the test time, we make prediction by (following Eq.~\ref{CLOMeq}),
\begin{align}
    \hat{y} = \argmax 
    \bigoplus_{1 \le k \le t} 
    \left[ \sigma_{k} f(h(\boldsymbol{x}, k)) + \mu_{k} \right]
    \label{calibratedpred}
\end{align}

\subsection{Protecting OOD Model of Each Task Using Masks}
\label{sec.hat}
We now discuss the task mask mechanism for protecting the OOD model of each task to deal with CF. 
In learning the OOD model for each task, CLOM at the same time also trains a \textit{mask} or \textit{hard attention} for each layer. To protect
{the shared feature extractor} 
from previous tasks, their masks are used to block those important neurons so that the new task learning will not interfere with the parameters learned for previous tasks. 

The main idea is to use sigmoid to approximate a 0-1 step function as \textit{hard attention} to mask (or block) or unmask (unblock) the information flow to protect the parameters learned for each previous task.

The hard attention (mask) at layer $l$ and task $t$ is defined as
\begin{align}
    \boldsymbol{a}_{l}^{t} = u\left( s \boldsymbol{e}_{l}^{t} \right) \label{attn}
\end{align}
where $u$ is the sigmoid function, $s$ is a scalar, and $\boldsymbol{e}_{l}^{t}$ is a \textit{learnable} embedding {of the task-id input $t$}.
The attention is element-wise multiplied to the output $\boldsymbol{h}_{l}$ of layer $l$ as
\begin{align}
    \boldsymbol{h}_{l}' = \boldsymbol{a}_{l}^{t} \otimes \boldsymbol{h}_{l}
\end{align}
{as depicted in Fig.~\ref{diagram}(b).} The sigmoid function $u$ converges to a 0-1 step function {as $s$ goes to infinity}. Since the true step function is not differentiable, a fairly large $s$ is chosen to achieve a differentiable pseudo step function based on sigmoid (see Appendix~\ref{hyperparam} for choice of $s$). The pseudo binary value of the attention determines how much information can flow forward and backward between adjacent layers.

Denote $\boldsymbol{h}_{l} = \text{ReLU}(\boldsymbol{W}_{l} \boldsymbol{h}_{l-1} + \boldsymbol{b}_{l})$, where $\text{ReLU}$ is the rectifier function. For units (neurons) of attention $\boldsymbol{a}_{l}^{t}$ with zero values, we can freely change the corresponding parameters in $\boldsymbol{W}_{l}$ and $\boldsymbol{b}_l$ without affecting the output $\boldsymbol{h}_{l}'$. For units of attention with non-zero values, changing the parameters will affect the output $\boldsymbol{h}_{l}'$ for which we need to protect from gradient flow in backpropagation to prevent forgetting.

Specifically, during training the new task $t$, we update parameters according to the attention so that the important parameters for
{past tasks ($1, ..., t-1$) are unmodified. Denote the accumulated attentions (masks) of all past tasks by 
\begin{align}
    \boldsymbol{a}_{l}^{<t} = \max ( \boldsymbol{a}_{l}^{< t-1}, \boldsymbol{a}_{l}^{t-1} )
\end{align}
where $\boldsymbol{a}_{l}^{<t}$ is the hard attentions of layer $l$ of all previous tasks, and $\max$ is an element-wise maximum\footnote{Some parameters from different tasks can be shared, which means some hard attention masks can be shared.} and $\boldsymbol{a}^{0}_{l}$ is a zero vector. Then the modified gradient is the following, 
\begin{align}
    \nabla w_{ij, l}' = \left( 1 - \min \left( a_{i, l}^{< t}, a_{j, l-1}^{< t} \right) \right) \nabla w_{ij, l} \label{hatupdate}
\end{align}
where $a^{< t}_{i, l}$ indicates $i$'th unit of $\boldsymbol{a}^{< t}_{l}$ and $l=1, ..., L-1$}.
This reduces the gradient if the corresponding units' attentions at layers $l$ and $l-1$ are non-zero\footnote{{By construction, if $\boldsymbol{a}_{l}^{t}$ becomes $\boldsymbol{1}$ for all layers, the gradients are zero and the network is at maximum capacity. However, the network capacity can increase by adding more parameters.}}.
We do not apply hard attention on the last layer $L$ because it is a task-specific layer. 

To encourage sparsity in $\boldsymbol{a}_{l}^{t}$ and parameter sharing with $\boldsymbol{a}_{l}^{<t}$, 
a regularization ($\mathcal{L}_{r}$) for attention at task $t$ is defined as 
\begin{align}
    \mathcal{L}_{r} = \lambda_{t} \frac{\sum_{l}\sum_{i} a_{i, l}^{t}\left( 1 - a_{i, l}^{< t} \right)}{\sum_{l}\sum_{i} \left( 1 - a_{i, l}^{< t} \right)}
\end{align}
where
$\lambda_t$ is a scalar hyperparameter.
For flexibility, we denote $\lambda_t$ for each task $t$. However, in practice, we use the same $\lambda_t$ for all $t \neq 1$.
The final objective to be minimized for task $t$ with hard attention is (see Fig.~\ref{diagram}(b))
\begin{align}
    \mathcal{L} = \mathcal{L}_{c} + \mathcal{L}_r \label{1obj}
\end{align}
where $\mathcal{L}_{c}$ is the contrastive loss function (Eq.~\ref{modsupclr}).
By protecting important parameters from changing during training,
the neural network effectively alleviates CF.

\section{Experiments} \label{experiment}
\textbf{Evaluation Datasets:} Four image classification CL benchmark datasets are used in our experiments.
(1) \textbf{MNIST}
:\footnote{http://yann.lecun.com/exdb/mnist/ 
} 
handwritten digits of 10 classes (digits) with 60,000 examples for training and 10,000 examples for testing.
(2) \textbf{CIFAR-10}~\citep{Krizhevsky2009learning}:\footnote{https://www.cs.toronto.edu/~kriz/cifar.html} 60,000 32x32 color images of 10 classes with 50,000 for training and 10,000 for testing. 
(3)~\textbf{CIFAR-100}~\citep{Krizhevsky2009learning}:\footnote{https://www.cs.toronto.edu/~kriz/cifar.html} 60,000 32x32 color images of 100 classes with 500 images per class for training and 100 per class for testing.~(4)~\textbf{Tiny-ImageNet}~\cite{Le2015TinyIV}:\footnote{http://tiny-imagenet.herokuapp.com} 120,000 64x64 color images of 200 classes with 500 images per class for training and 50 images per class for validation, and 50 images per class for testing. 
Since the test data has no labels in this dataset, we use the validation data as the test data as in \cite{Liu2020}.

\textbf{Baseline Systems:}
We compare our CLOM with both the classic and the most recent state-of-the-art 
CIL and TIL methods.
We also include CLOM(-c), which is CLOM without calibration (which already outperforms the baselines).

For CIL baselines, we consider seven \textit{replay} methods, \textbf{LwF.R} (replay version of LwF~\cite{Li2016LwF} with better results~\cite{Liu_2020_CVPR}), \textbf{iCaRL} \cite{Rebuffi2017}, \textbf{BiC}  \cite{wu2019large}, \textbf{A-RPS}~\cite{rajasegaran2020adaptive},
\textbf{Mnemonics} \cite{Liu_2020_CVPR}, \textbf{DER++} \cite{NEURIPS2020_b704ea2c_derpp}, and \textbf{Co$^2$L} \cite{Cha_2021_ICCV_co2l};
one {\textit{pseudo-replay}} method \textbf{CCG} \cite{abati2020conditional}\footnote{iTAML~\cite{rajasegaran2020itaml} is not included as they require a batch of test data from the same task to predict the task-id. When each batch has only one test sample, which is our setting, it is very weak. For example, iTAML TIL/CIL accuracy is only 35.2\%/33.5\% on CIFAR100 10 tasks. Expert Gate (EG)~\cite{Aljundi2016expert} is also weak. For example, its TIL/CIL accuracy is 87.2/43.2 on MNIST 5 tasks. Both iTAML and EG are much weaker than many baselines. };
one \textit{orthogonal projection} method
\textbf{OWM}~\cite{zeng2019continuous}; and a \textit{multi-classifier} method
\textbf{MUC}~\cite{Liu2020}; and \textit{a prototype augmentation method} \textbf{PASS}~\cite{Zhu_2021_CVPR_pass}. OWM, MUC, and PASS \textit{do not save any samples} from previous tasks.

TIL baselines include \textbf{HAT}~\cite{Serra2018overcoming}, \textbf{HyperNet}~\cite{von2019continual}, and \textbf{SupSup}~\cite{supsup2020}.
As noted earlier, CIL methods can also be used for TIL. In fact, TIL methods may be used for CIL too, but the results are very poor. We include them in our comparison

\subsection{Training Details}
\label{sec:traindetails}
For all experiments, we use 10\% 
of training data as the validation set to grid-search for good hyperparameters.
For minimizing the contrastive loss, we use LARS \cite{you2017large} for 700 epochs with initial learning rate 0.1. We linearly increase the learning rate by 0.1 per epoch for the first 10 epochs until 1.0 and then decay it by 
cosine scheduler \cite{loshchilov2016sgdr} after 10 epochs without restart as in~\cite{chen2020simple, tack2020csi}. For fine-tuning the classifier $f$, we use SGD for 100 epochs with learning rate 0.1 and reduce the learning rate by 0.1 at 60, 75, and 90 epochs. The full set of hyperparameters is given in Appendix~\ref{hyperparam}.

We follow the recent baselines (A-RPS, DER++, PASS and Co$^2$L) and use the same class split and backbone architecture for both CLOM and baselines. 

For MNIST and CIFAR-10, we split 10 classes into 5 tasks where each task has 2 classes in consecutive order. We save 20 random samples per class from the validation set for output calibration. This number is commonly used in replay methods~\cite{Rebuffi2017,wu2019large,rajasegaran2020adaptive,Liu_2020_CVPR}. MNIST consists of single channel images of size 1x28x28. Since the contrastive learning \cite{chen2020simple} relies on color changes, we copy the channel to make 3-channels. For MNIST and CIFAR-10, we use AlexNet-like architecture \cite{NIPS2012_c399862d} and ResNet-18 \cite{he2016deep} respectively for both CLOM and baselines.

For CIFAR-100, we conduct two experiments. We split 100 classes into 10 tasks and 20 tasks where each task has 10 and 5 classes, respectively, in consecutive order. We use 2000 memory budget as in \cite{Rebuffi2017}, saving 20 random samples per class from the validation set for output calibration. We use the same ResNet-18 structure for CLOM and baselines, but we increase the number of channels twice to learn more tasks.

For Tiny-ImageNet, we follow \cite{Liu2020} and resize the original images of size 3x64x64 to 3x32x32 so that the same ResNet-18 of CIFAR-100 experiment setting can be used. We split 200 classes into 5 tasks (40 classes per task) and 10 tasks (20 classes per task) in consecutive order, respectively.
To have the same memory budget of 2000 as for CIFAR-100, we save 10 random samples per class from the validation set for output calibration.

\textbf{Data Augmentation.} {For baselines, we use data augmentations used in their original papers. For CLOM, following \citep{chen2020simple, tack2020csi}, we use \textit{three initial augmentations} (see Sec.~\ref{unifiedappr}) (i.e., \textit{horizontal flip}, \textit{color change} (\textit{color jitter} and \textit{grayscale}), and \textit{Inception crop} \cite{inception}) and four \textit{rotations} (see Sec.~\ref{unifiedappr}). Specific details about these transformations are given in Appendix~\ref{aug_details}.}

\begin{table*}[t]
\centering
\resizebox{2.1\columnwidth}{!}{
\begin{tabular}{l c c c c c c c c c c c c | c c}
&&&&&&&&&&&&&&\\[-1.1em]
\toprule
\multirow{2}{*}{Method}  &  \multicolumn{2}{c}{MNIST-5T} & \multicolumn{2}{c}{CIFAR10-5T}  &  \multicolumn{2}{c}{CIFAR100-10T} &  \multicolumn{2}{c}{CIFAR100-20T} &  \multicolumn{2}{c}{T-ImageNet-5T} & \multicolumn{2}{c}{T-ImageNet-10T} & \multicolumn{2}{|c}{Average}\\
{} & TIL & CIL  &  TIL & CIL  &  TIL & CIL  &  TIL & CIL  &  TIL & CIL & TIL & CIL & TIL & CIL \\
&&&&&&&&&&&&&&\\[-1em]
\midrule
\multicolumn{13}{c}{\textbf{CIL Systems}} & \multicolumn{2}{|c}{} \\
\hline
OWM             &  99.7\scalebox{0.8}{$\pm$0.03}  & 95.8\scalebox{0.8}{$\pm$0.13}    & 85.2\scalebox{0.8}{$\pm$0.17}  & 51.7\scalebox{0.8}{$\pm$0.06}   & 59.9\scalebox{0.8}{$\pm$0.84}  & 29.0\scalebox{0.8}{$\pm$0.72}  & 65.4\scalebox{0.8}{$\pm$0.07}  & 24.2\scalebox{0.8}{$\pm$0.11} & 22.4\scalebox{0.8}{$\pm$0.87} & 10.0\scalebox{0.8}{$\pm$0.55} & 28.1\scalebox{0.8}{$\pm$0.55}  & 8.6\scalebox{0.8}{$\pm$0.42} & 60.1 & 36.5 \Tstrut \\
MUC         &  99.9\scalebox{0.8}{$\pm$0.02}  & 74.6\scalebox{0.8}{$\pm$0.45}    & 95.2\scalebox{0.8}{$\pm$0.24}  & 53.6\scalebox{0.8}{$\pm$0.95}   &  76.9\scalebox{0.8}{$\pm$1.27} & 30.0\scalebox{0.8}{$\pm$1.37}  & 73.7\scalebox{0.8}{$\pm$1.27}  & 14.4\scalebox{0.8}{$\pm$0.93} & 55.8\scalebox{0.8}{$\pm$0.26} & 33.6\scalebox{0.8}{$\pm$0.18}  & 47.2\scalebox{0.8}{$\pm$0.22}  & 17.4\scalebox{0.8}{$\pm$0.17} & 74.8 & 37.3 \\
PASS$^{\dagger}$&  99.5\scalebox{0.8}{$\pm$0.14}  & 76.6\scalebox{0.8}{$\pm$1.67}    & 83.8\scalebox{0.8}{$\pm$0.68}  & 47.3\scalebox{0.8}{$\pm$0.97}   &  72.4\scalebox{0.8}{$\pm$1.23} & 36.8\scalebox{0.8}{$\pm$1.64}  & 76.9\scalebox{0.8}{$\pm$0.77}  & 25.3\scalebox{0.8}{$\pm$0.81} & 50.3\scalebox{0.8}{$\pm$1.97} & 28.9\scalebox{0.8}{$\pm$1.36}  & 47.6\scalebox{0.8}{$\pm$0.38}  & 18.7\scalebox{0.8}{$\pm$0.58} & 71.8 & 38.9 \\ 
LwF.R         &  \textbf{99.9}\scalebox{0.8}{$\pm$0.09}  &   85.0\scalebox{0.8}{$\pm$3.05}  & 95.2\scalebox{0.8}{$\pm$0.30}  & 54.7\scalebox{0.8}{$\pm$1.18}   & 86.2\scalebox{0.8}{$\pm$1.00}  & 45.3\scalebox{0.8}{$\pm$0.75}  & 89.0\scalebox{0.8}{$\pm$0.45}  & 44.3\scalebox{0.8}{$\pm$0.46}  & 56.4\scalebox{0.8}{$\pm$0.48} & 32.2\scalebox{0.8}{$\pm$0.50} & 55.3\scalebox{0.8}{$\pm$0.35}  & 24.3\scalebox{0.8}{$\pm$0.26} & 80.3 & 47.6 \\
iCaRL$^*$         &  \textbf{99.9}\scalebox{0.8}{$\pm$0.09}  &  96.0\scalebox{0.8}{$\pm$0.42}   & 94.9\scalebox{0.8}{$\pm$0.34}  & 63.4\scalebox{0.8}{$\pm$1.11}   & 84.2\scalebox{0.8}{$\pm$1.04}  & 51.4\scalebox{0.8}{$\pm$0.99}  & 85.7\scalebox{0.8}{$\pm$0.68}  & 47.8\scalebox{0.8}{$\pm$0.48}   & 54.3\scalebox{0.8}{$\pm$0.59} & 37.0\scalebox{0.8}{$\pm$0.41} & 52.7\scalebox{0.8}{$\pm$0.37}  & 28.3\scalebox{0.8}{$\pm$0.18} & 78.6 & 54.0 \\
Mnemonics$^{\dagger *}$ &  \textbf{99.9}\scalebox{0.8}{$\pm$0.03}  &  96.3\scalebox{0.8}{$\pm$0.36}   & 94.5\scalebox{0.8}{$\pm$0.46}  & 64.1\scalebox{0.8}{$\pm$1.47}   & 82.3\scalebox{0.8}{$\pm$0.30}  & 51.0\scalebox{0.8}{$\pm$0.34}   & 86.2\scalebox{0.8}{$\pm$0.46}  & 47.6\scalebox{0.8}{$\pm$0.74}  & 54.8\scalebox{0.8}{$\pm$0.16} & 37.1\scalebox{0.8}{$\pm$0.46} & 52.9\scalebox{0.8}{$\pm$0.66}  & 28.5\scalebox{0.8}{$\pm$0.72} & 78.5 & 54.1 \\ 
BiC   &  \textbf{99.9}\scalebox{0.8}{$\pm$0.04}  &  85.1\scalebox{0.8}{$\pm$1.84}   & 91.1\scalebox{0.8}{$\pm$0.82}  & 57.1\scalebox{0.8}{$\pm$1.09}   & 87.6\scalebox{0.8}{$\pm$0.28}  & 51.3\scalebox{0.8}{$\pm$0.59}   & 90.3\scalebox{0.8}{$\pm$0.26}  &  40.1\scalebox{0.8}{$\pm$0.77}  & 44.7\scalebox{0.8}{$\pm$0.71} & 20.2\scalebox{0.8}{$\pm$0.31} & 50.3\scalebox{0.8}{$\pm$0.65}  & 21.2\scalebox{0.8}{$\pm$0.46} & 77.3 & 45.8 \\
DER++ & 99.7\scalebox{0.8}{$\pm$0.08} &    95.3\scalebox{0.8}{$\pm$0.69} & 92.2\scalebox{0.8}{$\pm$0.48}  & 66.0\scalebox{0.8}{$\pm$1.27}  & 84.2\scalebox{0.8}{$\pm$0.47} &  55.3\scalebox{0.8}{$\pm$0.10}  & 86.6\scalebox{0.8}{$\pm$0.50}  & 46.6\scalebox{0.8}{$\pm$1.44} & 58.0\scalebox{0.8}{$\pm$0.52} & 36.0\scalebox{0.8}{$\pm$0.42} & 59.7\scalebox{0.8}{$\pm$0.6} & 30.5\scalebox{0.8}{$\pm$0.30} & 80.1 & 55.0 \\
A-RPS &      &       &     &      &     & 60.8  &  & 53.5  &   & & & \\
CCG  &      & \textbf{97.3}  &  & 70.1   &   &  &  &  &  & & & \\
Co$^2$L     &  &   & 93.4 & 65.6 &  &  &  &  &  & & & \\
\hline
\multicolumn{13}{c}{\textbf{TIL Systems}} & \multicolumn{2}{|c}{} \Tstrut\Bstrut \\ 
\hline
HAT             &  \textbf{99.9}\scalebox{0.8}{$\pm$0.02}  &  81.9\scalebox{0.8}{$\pm$3.73}   & 96.7\scalebox{0.8}{$\pm$0.18}  & 62.7\scalebox{0.8}{$\pm$1.46}   & 84.0\scalebox{0.8}{$\pm$0.23}  & 41.1\scalebox{0.8}{$\pm$0.93}   & 85.0\scalebox{0.8}{$\pm$0.85}  & 26.0\scalebox{0.8}{$\pm$0.83}  & 61.2\scalebox{0.8}{$\pm$0.72} & 38.5\scalebox{0.8}{$\pm$1.85} & 63.8\scalebox{0.8}{$\pm$0.41}  & 29.8\scalebox{0.8}{$\pm$0.65} & 81.8 & 46.6 \Tstrut \\
HyperNet        &  99.7\scalebox{0.8}{$\pm$0.05} &  49.1\scalebox{0.8}{$\pm$5.52}   & 94.9\scalebox{0.8}{$\pm$0.54}  & 47.4\scalebox{0.8}{$\pm$5.78}   & 77.3\scalebox{0.8}{$\pm$0.45}  & 29.7\scalebox{0.8}{$\pm$2.19}   & 83.0\scalebox{0.8}{$\pm$0.60}  & 19.4\scalebox{0.8}{$\pm$1.44} & 23.8\scalebox{0.8}{$\pm$1.21} & 8.8\scalebox{0.8}{$\pm$0.98} & 27.8\scalebox{0.8}{$\pm$0.86}  & 5.8\scalebox{0.8}{$\pm$0.56} & 67.8 & 26.7 \\
SupSup & 99.6\scalebox{0.8}{$\pm$0.09} &    19.5\scalebox{0.8}{$\pm$0.15} & 95.3\scalebox{0.8}{$\pm$0.27}  & 26.2\scalebox{0.8}{$\pm$0.46}  & 85.2\scalebox{0.8}{$\pm$0.25} &  33.1\scalebox{0.8}{$\pm$0.47}  & 88.8\scalebox{0.8}{$\pm$0.18}  & 12.3\scalebox{0.8}{$\pm$0.30} & 61.0\scalebox{0.8}{$\pm$0.62} & 36.9\scalebox{0.8}{$\pm$0.57} & 64.4\scalebox{0.8}{$\pm$0.20} & 27.0\scalebox{0.8}{$\pm$0.45} & 82.4 & 25.8  \\
\hline
CLOM(-c)             &  \textbf{99.9}\scalebox{0.8}{$\pm$0.00}  &  94.4\scalebox{0.8}{$\pm$0.26}   & \textbf{98.7}\scalebox{0.8}{$\pm$0.06}  & 87.8\scalebox{0.8}{$\pm$0.71}   & \textbf{92.0}\scalebox{0.8}{$\pm$0.37}  & 63.3\scalebox{0.8}{$\pm$1.00}   & \textbf{94.3}\scalebox{0.8}{$\pm$0.06}  & 54.6\scalebox{0.8}{$\pm$0.92}  & \textbf{68.4}\scalebox{0.8}{$\pm$0.16} & 45.7\scalebox{0.8}{$\pm$0.26} & \textbf{72.4}\scalebox{0.8}{$\pm$0.21}  & 47.1\scalebox{0.8}{$\pm$0.18} & 87.6 & 65.5 \Tstrut \\ 
CLOM         &  \textbf{99.9}\scalebox{0.8}{$\pm$0.00}  &  96.9\scalebox{0.8}{$\pm$0.30}   & \textbf{98.7}\scalebox{0.8}{$\pm$0.06}  & \textbf{88.0}\scalebox{0.8}{$\pm$0.48}   & \textbf{92.0}\scalebox{0.8}{$\pm$0.37}  & \textbf{65.2}\scalebox{0.8}{$\pm$0.71}   & \textbf{94.3}\scalebox{0.8}{$\pm$0.06}  & \textbf{58.0}\scalebox{0.8}{$\pm$0.45} & \textbf{68.4}\scalebox{0.8}{$\pm$0.16} & \textbf{51.7}\scalebox{0.8}{$\pm$0.37} & \textbf{72.4}\scalebox{0.8}{$\pm$0.21}  & \textbf{47.6}\scalebox{0.8}{$\pm$0.32} & \textbf{87.6} & \textbf{67.9} \\
\bottomrule
\end{tabular}
}
\caption{
Average accuracy over all classes after the last task is learned. -xT: x number of tasks.  {$\dagger$: In their original paper, PASS and Mnemonics use the first half of classes to pre-train before CL. Their results are 50.1\% and 53.5\% on CIFAR100-10T respectively, but they are still lower than CLOM without pre-training. In our experiments, no pre-training is used  for fairness.} $^{*}$: \textbf{iCaRL} and \textbf{Mnemonics} give both the final average accuracy as here and the \textit{average incremental accuracy} in the original papers. We report the \textit{average incremental accuracy} and \textit{network size} in Appendix~\ref{avg_inc_acc} and \ref{param_size}, respectively. The last two columns show the average TIL and CIL accuracy of each method over all datasets.
}
\label{Tab:maintable}
\end{table*}

\subsection{Results and Comparative Analysis} \label{resultcomparison}
As in existing works, we evaluate each method by two metrics: \textit{average classification accuracy} on all classes after training the last task, and \textit{average forgetting rate}~\cite{Liu_2020_CVPR}, $F^{t} = \frac{1}{t-1}\sum_{j=1}^{t-1} A_{j}^{\text{init}} - A_{j}^{t}$, 
where $A_{j}^{\text{init}}$ is the $j$'th task's accuracy of the network right after the $j$'th task is learned and $A_{j}^{t}$ is the accuracy of the network on the $j$'th task data after learning the last task $t$. We report the forgetting rate after the final task $t$. Our results are averages of 5 random runs.

We present the main experiment results in Tab.~\ref{Tab:maintable}. The last two columns give the average TIL/CIL results of each system/row.  For A-RPS, CCG, and Co$^2$L, we copy the results from their original papers as their codes are not released to the public or the public code cannot run on our system. The rows are grouped by CIL and TIL methods. 

\textbf{CIL Results Comparison}. {Tab. \ref{Tab:maintable} shows that CLOM and CLOM(-c) achieve much higher CIL accuracy except for MNIST for which CLOM is slightly weaker than CCG by 0.4\%, but CLOM's result on CIFAR10-5T is about 18\% greater than CCG.}
For other datasets, CLOM improves by similar margins. {This is in contrast to the baseline TIL systems that are incompetent at the CIL setting when classes are predicted using Eq. \ref{CLOMeq}.}
Even \textit{without} calibration,
CLOM(-c) already outperforms all the baselines by large margins.

\textbf{TIL Results Comparison}.
{The gains by CLOM and CLOM(-c) over the baselines are also great in the TIL setting.}
CLOM and CLOM(-c) are the same as the output calibration does not affect TIL performance. For the two large datasets CIFAR100 and T-ImageNet, CLOM gains by large margins. {
This is due to contrastive learning and the OOD model. The replay based CIL methods (LwF.R, iCaRL, Mnemonics, BiC, and DER++) perform reasonably in the TIL setting, but our CLOM and CLOM(-c) are much better due to task masks which can protect previous models better with little CF.}

\begin{figure}[!ht]
    \includegraphics[width=3.3in]{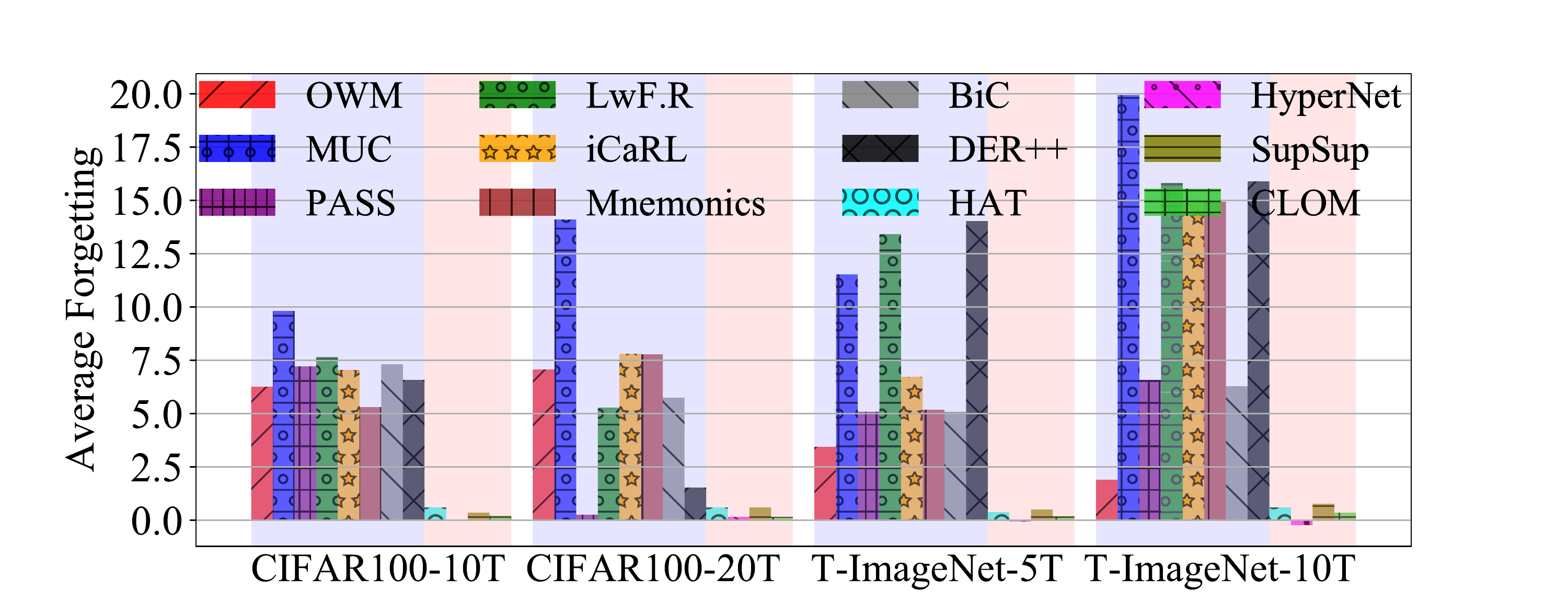}
    \caption{Average forgetting rate (\%) in the TIL setting as CLOM is a TIL system.
    The lower the value, the better the method is. {CIL/TIL systems are shaded in blue/red, respectively (\textit{best viewed in color}).} A negative value indicates the task accuracy has increased from the initial accuracy.} 
    \label{forgetplot}
\end{figure}

\textbf{Comparison of Forgetting Rate}. Fig. \ref{forgetplot} shows the average forgetting rate of each method in the TIL setting. The CIL systems suffer from more forgetting as they are not designed for the TIL setting, which results in lower TIL accuracy (Tab. \ref{Tab:maintable}). 
The TIL systems are highly {effective at preserving previous within-task knowledge}. This results in higher TIL accuracy on large dataset such as T-ImageNet, but they collapse when task-id is not provided (the CIL setting)
as shown in Tab. \ref{Tab:maintable}. 
{CLOM is robust to forgetting as a TIL system and it also functions well without task-id.}

We report only the forgetting rate in the TIL setting because our CLOM is essentially a TIL method and not a CIL system by design. The degrading CIL accuracy of CLOM is mainly because the OOD model for each task is not perfect.

\subsection{Ablation Studies} \label{ablation}

\begin{table}
\centering
\resizebox{\columnwidth}{!}{
\begin{tabular}{lcccccccc}
\toprule
                  & \multicolumn{4}{c}{CIFAR10-5T} & \multicolumn{4}{c}{CIFAR100-10T} \\
                  & \multicolumn{1}{l}{AUC} & \multicolumn{1}{l}{TaskDR}    & \multicolumn{1}{c}{TIL}       & \multicolumn{1}{l}{CIL} & \multicolumn{1}{l}{AUC} & \multicolumn{1}{l}{TaskDR}   & \multicolumn{1}{c}{TIL} & \multicolumn{1}{l}{CIL} \\
\midrule
SupSup           & 78.9 & 26.2 & 95.3 & 26.2 & 76.7                      & 34.3          & 85.2 & 33.1                    \\
SupSup (OOD in CLOM)       & 88.9 & 82.3 & 97.2 & 81.5 & 84.9                      & 63.7          & 90.0 & 62.1                    \\
\midrule
CLOM (ODIN)      & 82.9 & 63.3 & 96.7 & 62.9  & 77.9                      & 43.0          & 84.0 & 41.3                    \\
CLOM             & 92.2 & 88.5 & 98.7 & 88.0  & 85.0                      & 66.8           & 92.0 & 65.2    \\
\midrule
CLOM (w/o OOD) & 90.3 & 83.9 & 98.1 & 83.3 & 82.6                      & 59.5           & 89.8 & 57.5                    \\
\bottomrule

\end{tabular}
}
\caption{{TIL and CIL results improve with better OOD detection. Column AUC gives the average AUC score for the OOD detection method as used within each system on the left. Column TaskDR gives \textit{task detection rate}. TIL and CIL results are average accuracy values. {SupSup and CLOM variants are calibrated with 20 samples per class.}}
}
\label{betterood}
\end{table}

{\textbf{Better OOD for better continual learning.} We show that (1) an existing CL model can be improved by a good OOD model and (2) CLOM's results will deteriorate if  a weaker OOD model is applied. 
To isolate effect of OOD detection on changes in CIL performance, we further define \textit{task detection} and \textit{task detection rate}. For a test sample from a class of task $j$, if it is predicted to a class of task $m$ and $j=m$, the task detection is correct for this test instance. The \textit{task detection rate} $\sum_{\mathbf{x} \in \mathcal{D}^{\text{test}} } 1_{j = m}/N$, where $N$ is the total number of test instances in $\mathcal{D}^{\text{test}}$, is the rate of correct task detection. We measure the performance of OOD detection using AUC (Area Under the ROC Curve) averaged over all tasks. AUC is the main measure used in OOD detection papers. We conduct experiments on CIFAR10-5T and CIFAR100-10T.

For (1), we use the TIL baseline SupSup as it displays a strong TIL performance and is robust to forgetting like CLOM. We replace SupSup’s task learner with the OOD model in CLOM. Tab.~\ref{betterood} shows that the OOD method in CLOM improves SupSup (SupSup (OOD in CLOM)) greatly. It shows that our approach is applicable to different TIL systems.

For (2), we replace CLOM’s OOD method with a weaker OOD method ODIN~\cite{liang2018enhancing}. We see in Tab.~\ref{betterood} that task detection rate, TIL, and CIL results all drop markedly with ODIN (CLOM (ODIN)).
}

{\textbf{CLOM without OOD detection.} In this case, CLOM uses contrastive learning and data augmentation,
but does not use the rotation classes in classification. Note that the rotation classes are basically regarded as OOD data in training and for OOD detection in testing. CLOM (w/o OOD) in Tab.~\ref{betterood} represents this CLOM variant. We see that CLOM (w/o OOD) is much weaker than the full CLOM. This indicates that the improved results of CLOM over baselines are not only due to contrastive learning and data augmentation but also significantly due to OOD detection.}

\begin{table}
\parbox{.50\linewidth}{
\centering
\resizebox{.50\columnwidth}{!}{
\begin{tabular}{lcc}
\toprule
\multicolumn{1}{c}{$|\mathcal{M}|$} & (a) & (b) \\
\midrule
0 & 63.3\scalebox{0.8}{$\pm$1.00} & 54.6\scalebox{0.8}{$\pm$0.92} \\
5 & 64.9\scalebox{0.8}{$\pm$0.67} & 57.7\scalebox{0.8}{$\pm$0.50} \\
10 & 65.0\scalebox{0.8}{$\pm$0.71} & 57.8\scalebox{0.8}{$\pm$0.53} \\
15 & 65.1\scalebox{0.8}{$\pm$0.71} & 57.9\scalebox{0.8}{$\pm$0.44} \\
20 & 65.2\scalebox{0.8}{$\pm$0.71} & 58.0\scalebox{0.8}{$\pm$0.45} \\
\bottomrule
\end{tabular}
}
}
\hfill
\parbox{.46\linewidth}{
\centering
\resizebox{.46\columnwidth}{!}{
\begin{tabular}{lccc}
\toprule
   s        & $F^{5}$          & AUC     & CIL     \\
\midrule
1   & 48.6 & 58.8 & 10.0 \\
100 & 13.3 & 82.7 & 67.7 \\
300 & 8.2  & 83.3 & 72.0 \\
500 & 0.2 & 91.8 & 87.2 \\
700 & 0.1 & 92.2 & 88.0 \\
\bottomrule
\end{tabular}
}
}
\caption{(\textit{Left}) shows changes of accuracy with the number of samples saved per class for output calibration. (a) and (b) are CIFAR100-10T and CIFAR100-20T, respectively. $|\mathcal{M}|=k$ indicates $k$ samples are saved per class. {(\textit{Right}) shows that weaker forgetting mechanism results in larger forgetting and lower AUC, thus lower CIL. For $s=1$, the pseudo-step function becomes the standard sigmoid, thus parameters are hardly protected. $F^{5}$ is the forgetting rate over 5 tasks.}}
\label{memory_and_s}
\end{table}

\textbf{Effect of the number of saved samples for calibration.}
Tab. \ref{memory_and_s} (left) reveals that the output calibration is still effective with a small number of saved samples per class ($|\mathcal{M}|$). For both CIFAR100-10T and CIFAR100-20T, {CLOM achieves competitive performance by using only 5 samples per class.} The accuracy improves and the variance decreases with the number of saved samples.

{\textbf{Effect of $s$ in Eq.~\ref{attn} on forgetting of CLOM.}
We need to use a strong forgetting mechanism for CLOM to be functional. Using CIFAR10-5T, we show how CLOM performs with different values of $s$ in hard attention or masking. The larger $s$ value, the stronger protection is used. Tab.~\ref{memory_and_s} (right) shows that average AUC and CIL decrease as the forgetting rate increases. This also supports the result in Tab.~\ref{betterood} that SupSup improves greatly with the OOD method in CLOM as it is also robust to forgetting. PASS and Co$^2$L underperform despite they also use rotation or constrastive loss as their forgetting mechanisms are weak.
}

\begin{table}
\centering
\resizebox{\columnwidth}{!}{
\begin{tabular}{lcccc}
\toprule
\multicolumn{1}{}{} & \multicolumn{2}{c}{CIFAR10-5T} & \multicolumn{2}{c}{CIFAR100-10T} \\
   Aug.        & TIL          & CIL     & TIL & CIL     \\
\midrule
Hflip      & (93.1, 95.3) & (49.1, 72.7) & (77.6, 84.0) & (31.1, 47.0) \\
Color      & (91.7, 94.6) & (50.9, 70.2) & (67.2, 77.4) & (28.7, 41.8) \\
Crop       & (96.1, 97.3) & (58.4, 79.4) & (84.1, 89.3) & (41.6, 60.3) \\
All & (97.6, 98.7) & (74.0, 88.0) & (88.1, 92.0) & (50.2, 65.2) \\
\bottomrule
\end{tabular}
}
\caption{Accuracy of CLOM variants when a single or all augmentations are applied.
Hflip: horizontal flip; Color: color jitter and grayscale; Crop: Inception~\cite{inception}. (num1, num2): accuracy without and with rotation.
}
\label{augmentation}
\end{table}

\textbf{Effect of data augmentations.} 
{For data augmentation, we use three initial augmentations (i.e., \textit{horizontal flip}, \textit{color change} (color jitter and grayscale), \textit{Inception crop} \cite{inception}), which are commonly used in contrastive learning to build a single model. We additionally use rotation for OOD data in training. To evaluate the contribution of each augmentation
when task models are trained sequentially,
we train CLOM using one augmentation. We do not report their effects on forgetting as we experience rarely any forgetting (Fig. \ref{forgetplot} and Tab. \ref{memory_and_s}). Tab. \ref{augmentation} shows that the performance is lower when only a single augmentation is applied.
When all augmentations are applied, the TIL/CIL accuracies are higher. The rotation always improves the result when it is combined with other augmentations.
More importantly, when we use crop and rotation, we achieve higher CIL accuracy (79.4/60.3\% for CIFAR10-5T/CIFAR100-10T) than we use all augmentations without rotation (74.0/50.2\%).
This shows the efficacy of rotation in our system.}

\section{Conclusions} 

This paper proposed a novel continual learning (CL) method called CLOM based on OOD detection and task masking that can perform both task incremental learning (TIL) and class incremental learning (CIL). Regardless whether it is used for TIL or CIL in testing, the training process is the same, which is an advantage over existing CL systems as they focus on either CIL or TIL and have limitations on the other problem. Experimental results showed that CLOM outperforms both state-of-the-art TIL and CIL methods by very large margins.
In our future work, we will study ways to improve efficiency and also accuracy.

\section*{Acknowledgments}
{\color{black}Gyuhak Kim, Sepideh Esmaeilpour and Bing Liu were supported in part by two National Science Foundation (NSF) grants (IIS-1910424 and IIS-1838770), a DARPA Contract HR001120C0023, a KDDI research contract, and a Northrop Grumman research gift.} 

\bibliography{clom22}
\appendix
\section{Average Incremental Accuracy}\label{avg_inc_acc}
\begin{figure*}
\centering     
\subfigure{\includegraphics[width=68mm]{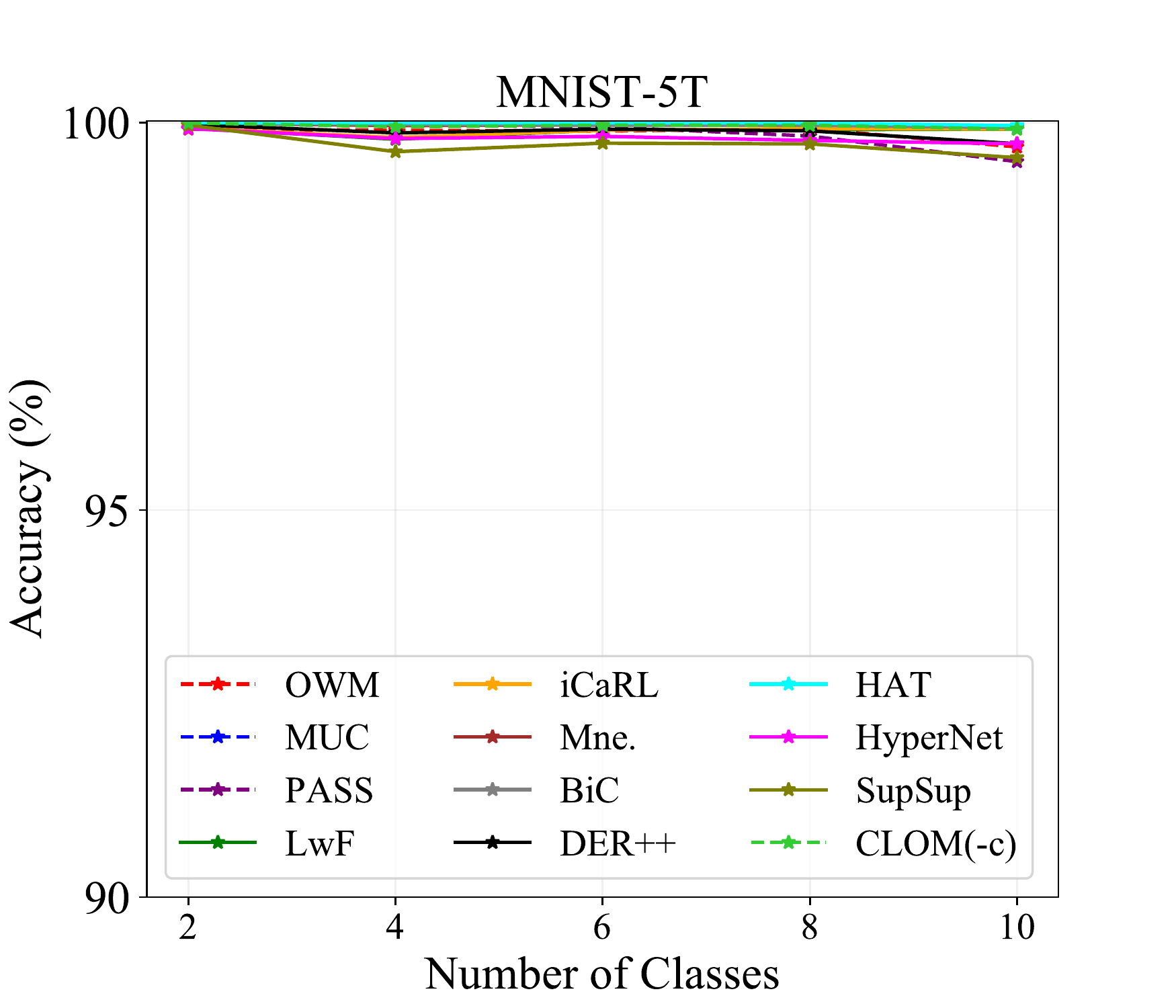}}
\subfigure{\includegraphics[width=68mm]{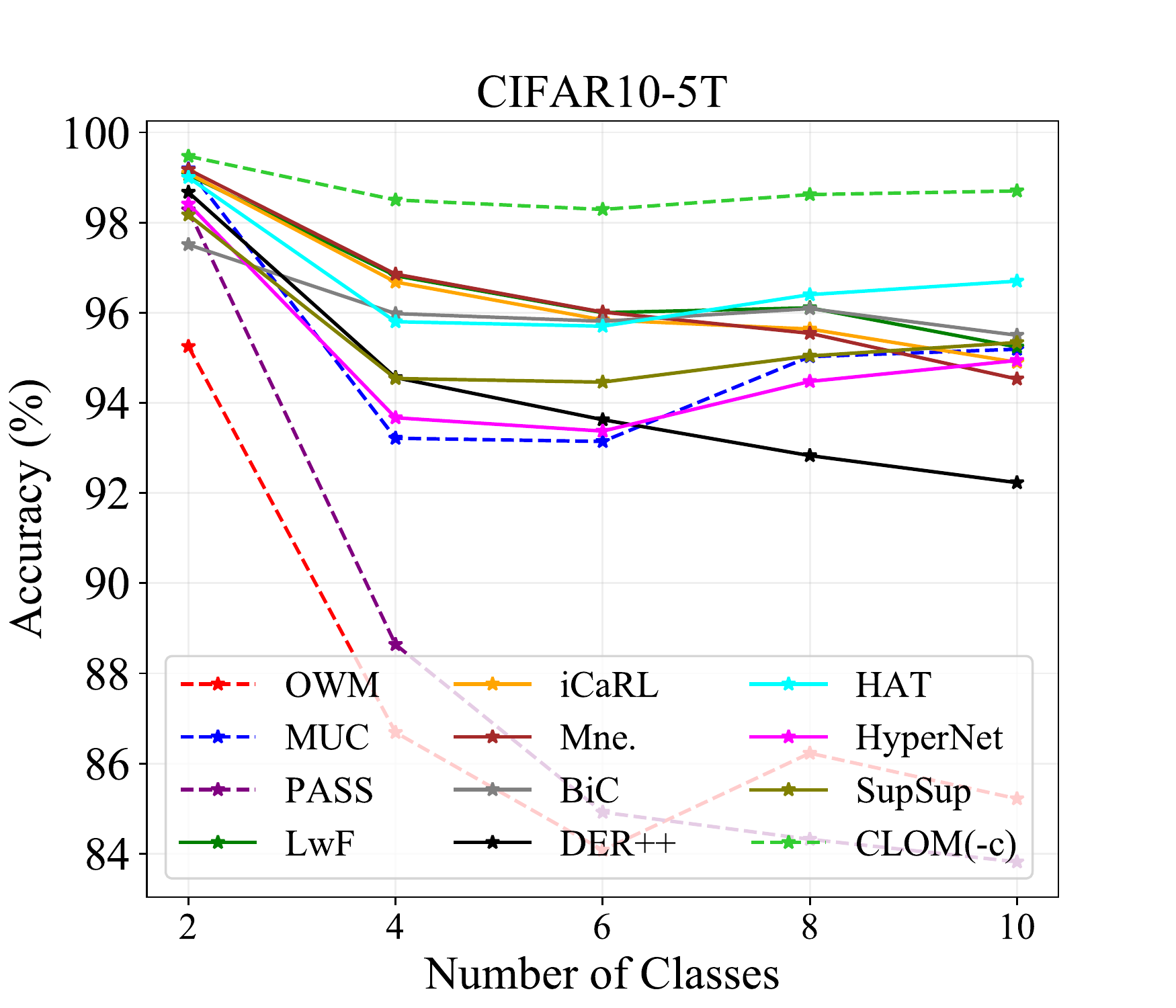}}
\subfigure{\includegraphics[width=68mm]{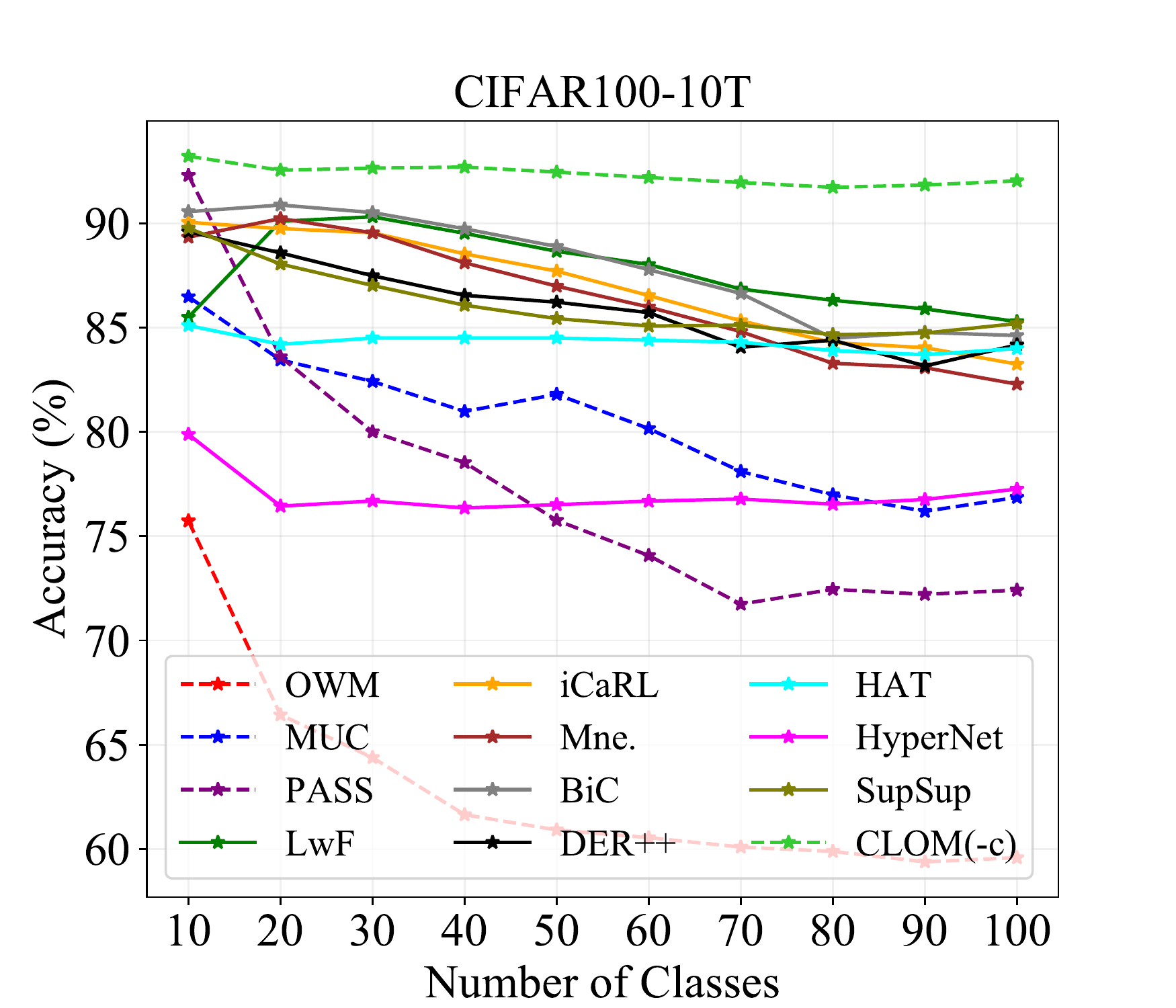}}
\subfigure{\includegraphics[width=68mm]{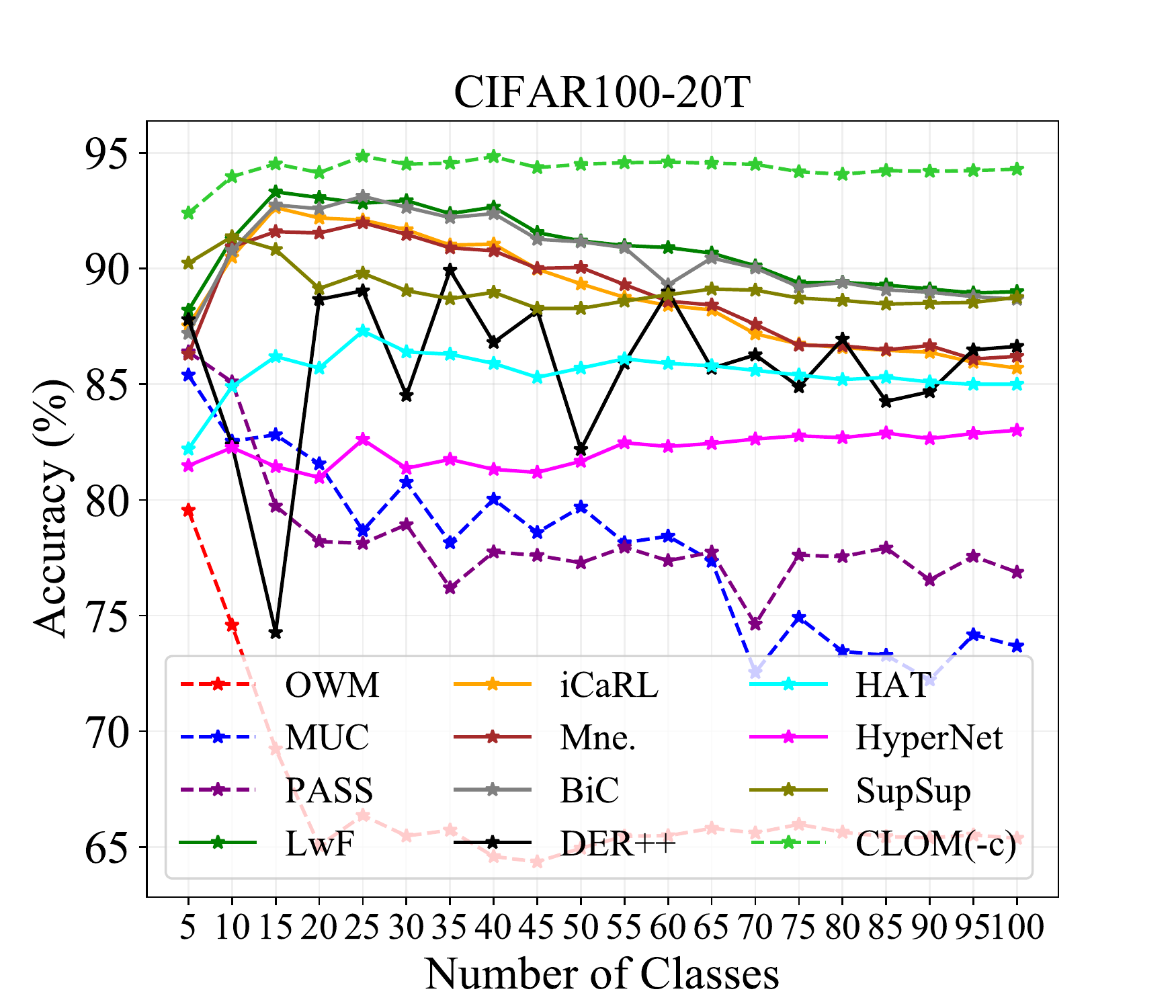}}
\subfigure{\includegraphics[width=68mm]{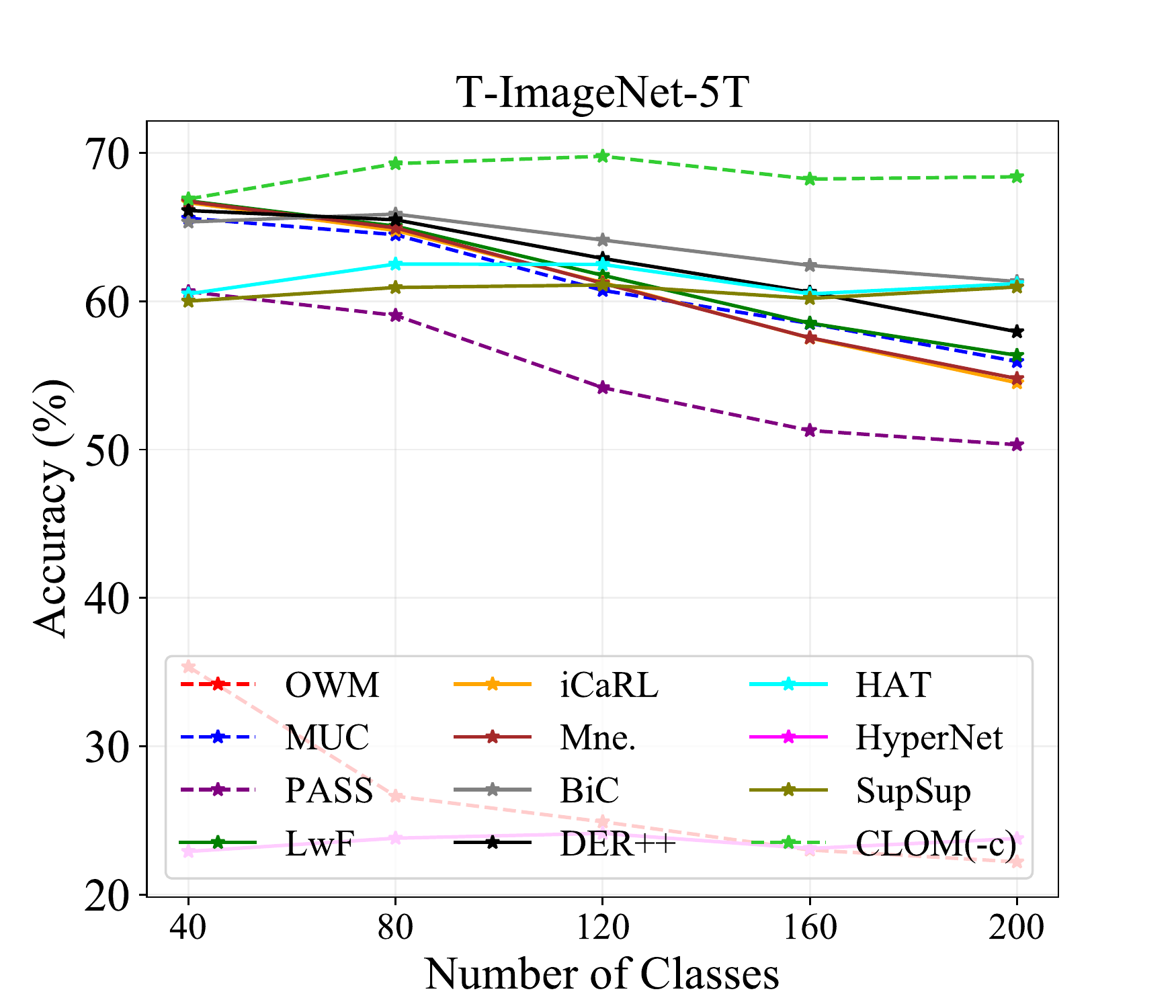}}
\subfigure{\includegraphics[width=68mm]{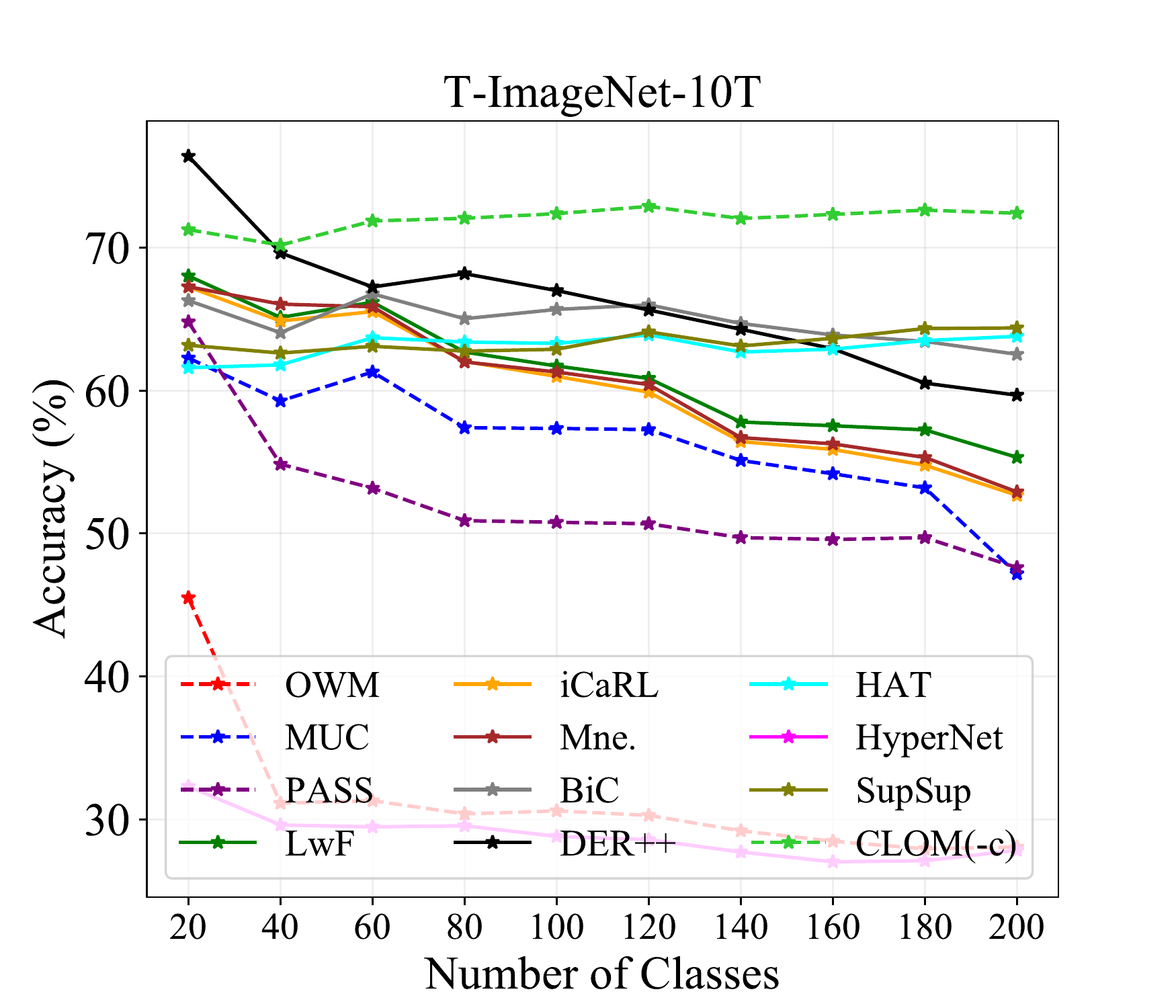}}
\caption{TIL performance over number of classes. The dashed lines indicate the methods that do not save any samples from previous tasks. The calibrated version, CLOM, is omitted as its TIL accuracy is the same as CLOM(-c). Best viewed in color.
}
\label{tilcurve}
\end{figure*}

\begin{figure*}
\centering     
\subfigure{\includegraphics[width=68mm]{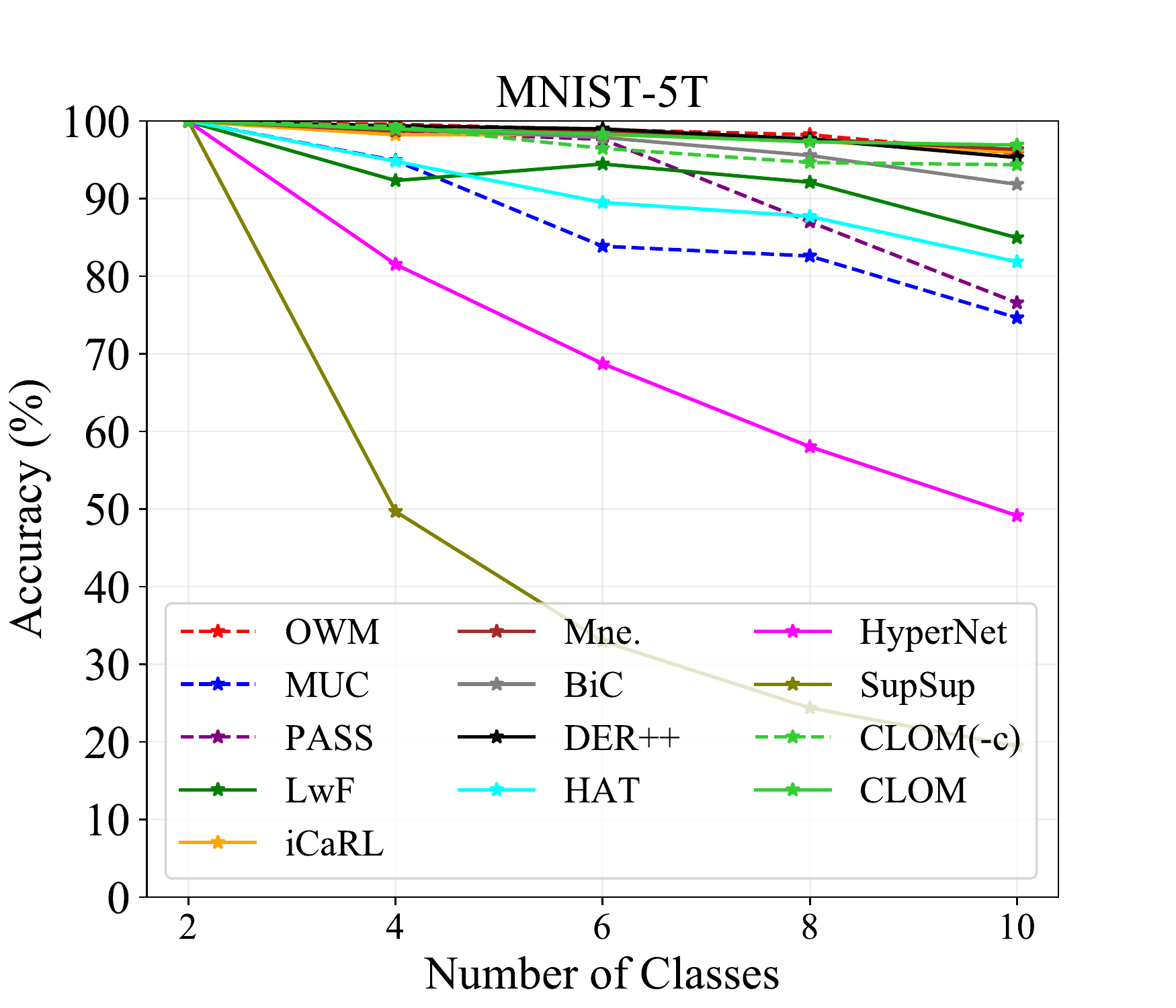}}
\subfigure{\includegraphics[width=68mm]{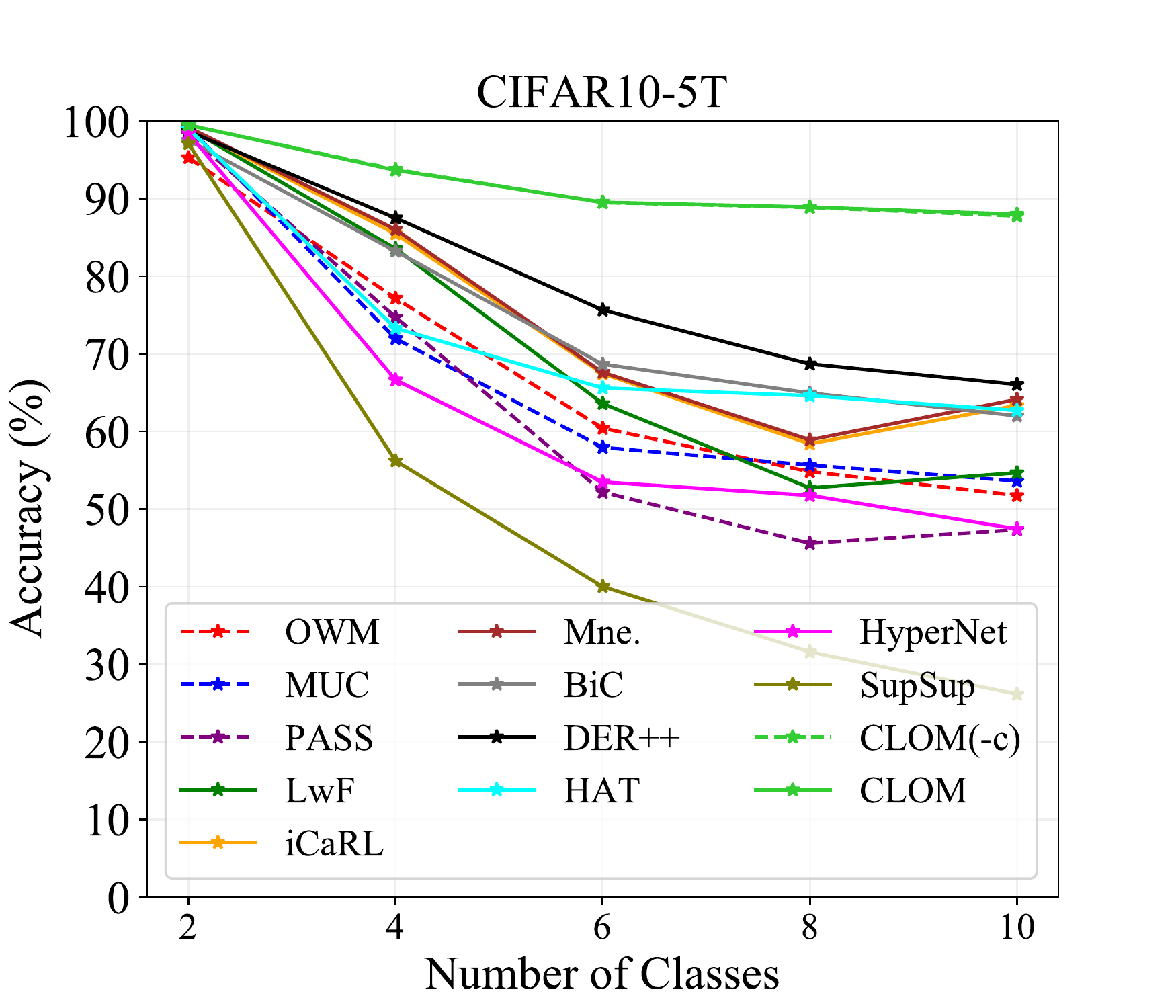}}
\subfigure{\includegraphics[width=68mm]{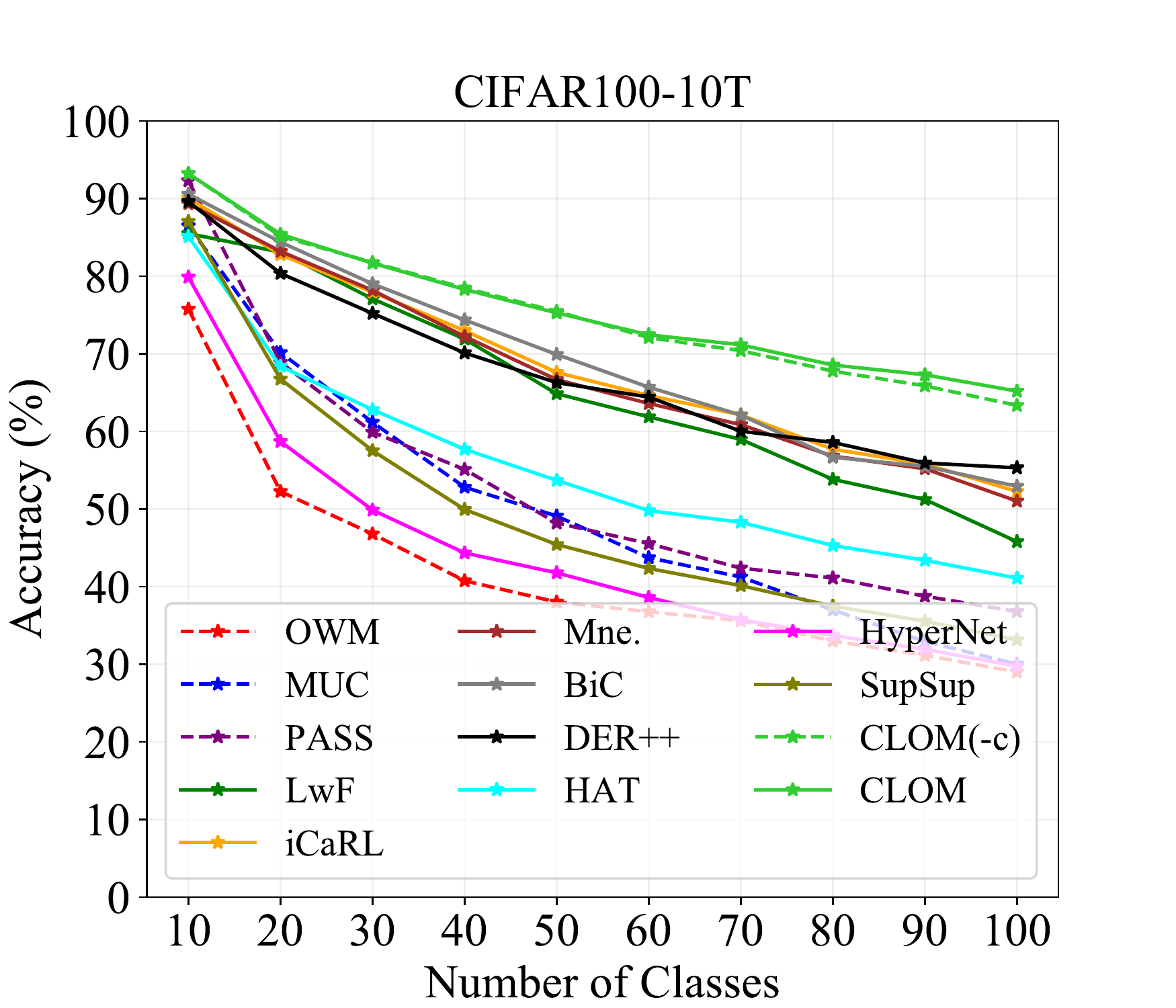}}
\subfigure{\includegraphics[width=68mm]{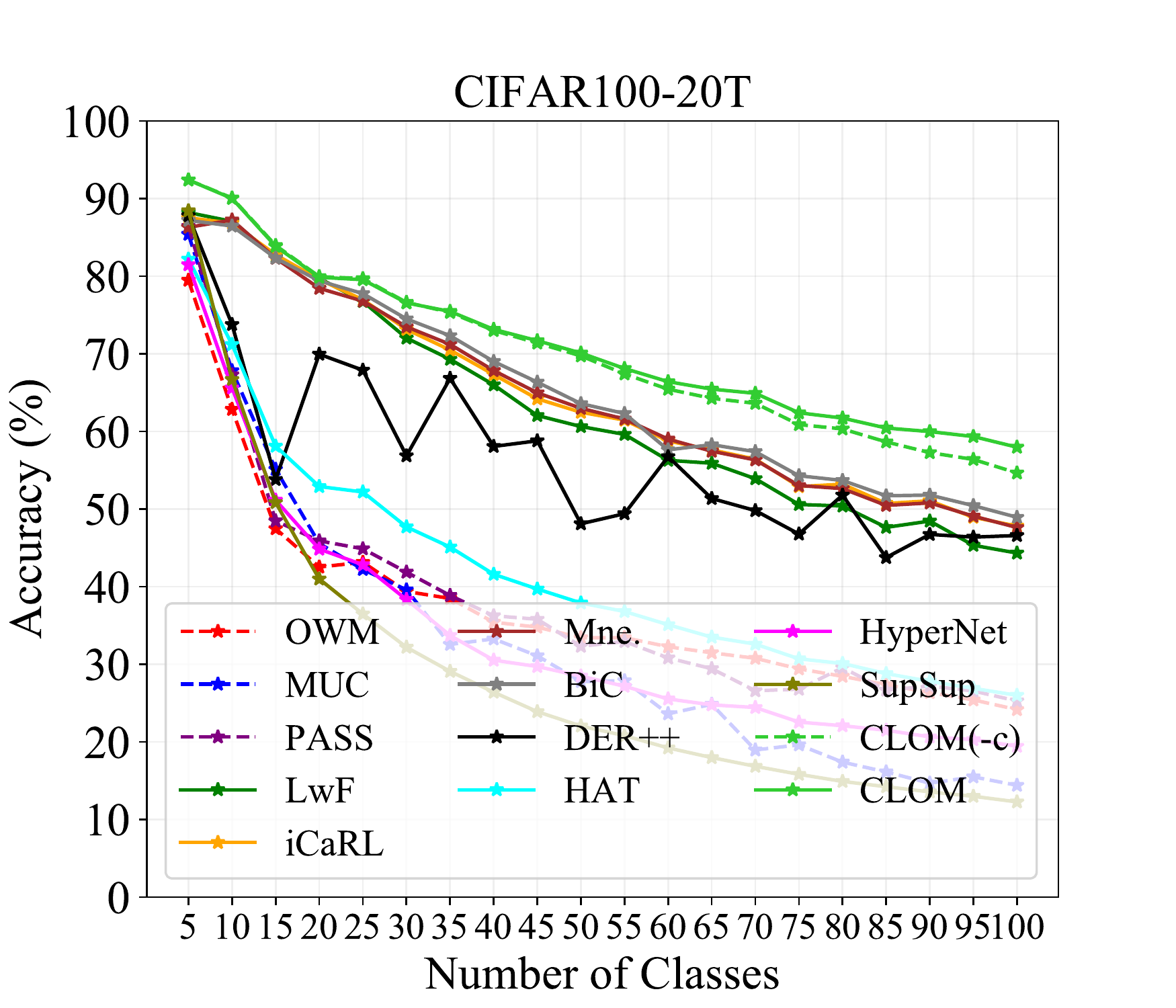}}
\subfigure{\includegraphics[width=68mm]{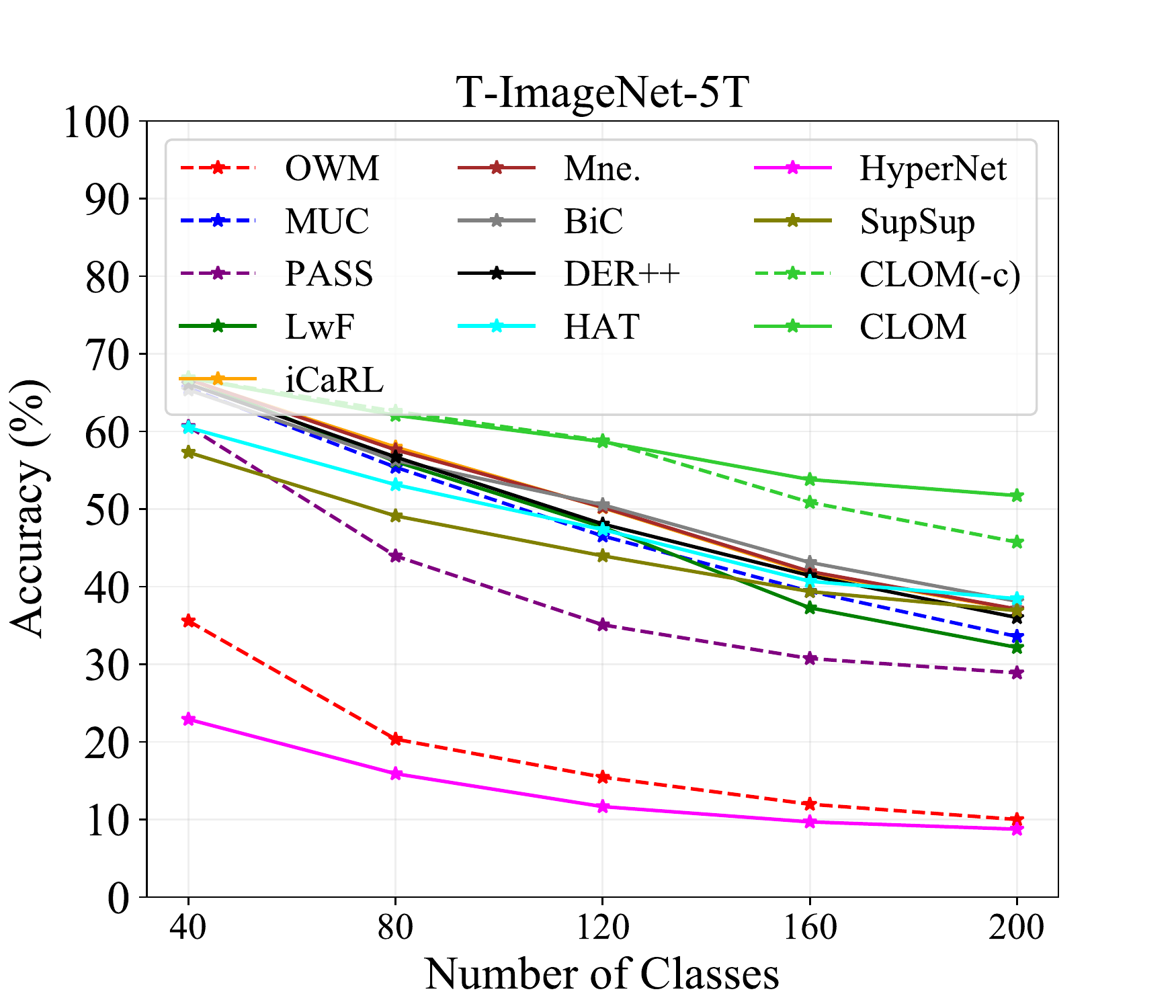}}
\subfigure{\includegraphics[width=68mm]{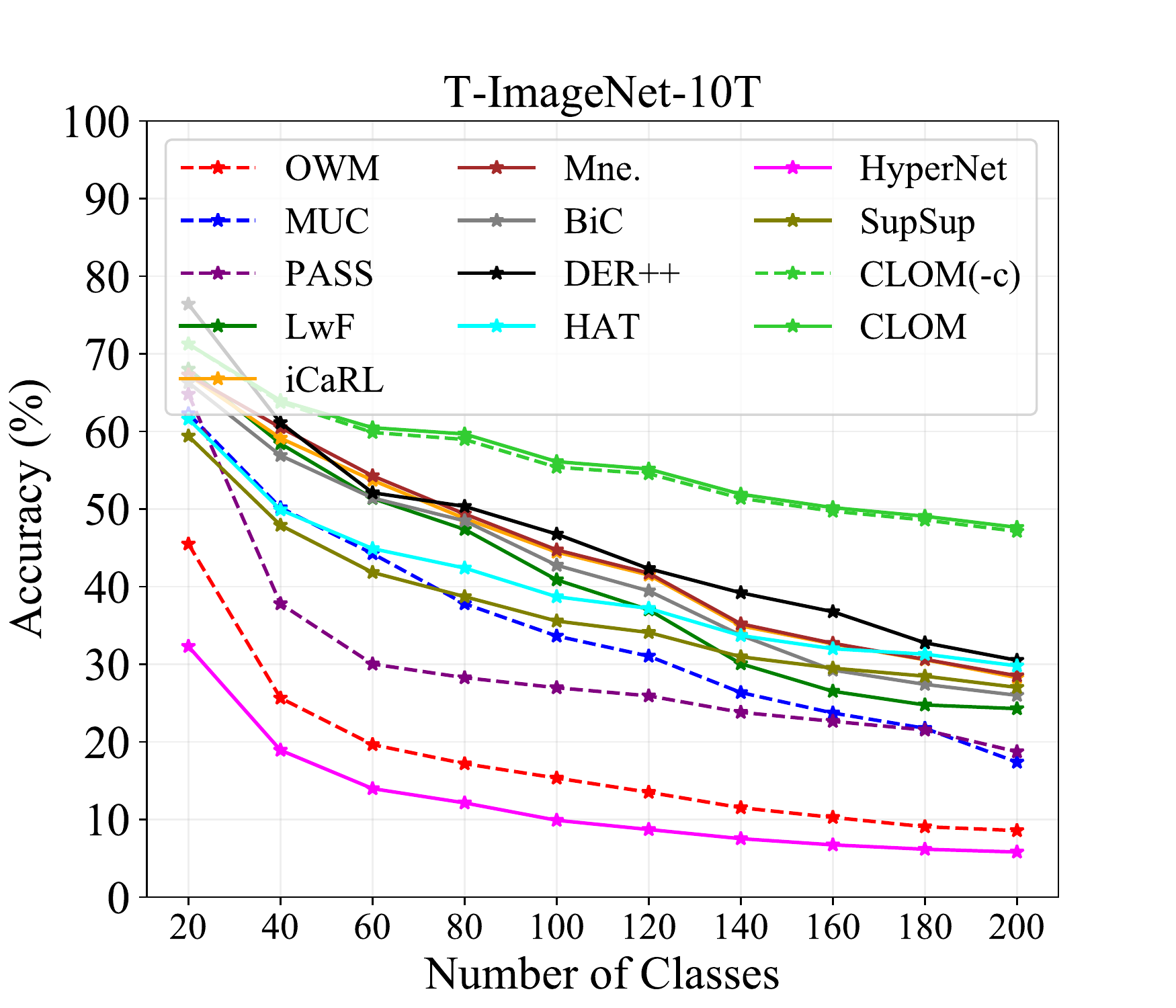}}
\caption{CIL performance over number of classes. The dashed lines indicate the methods that do not save any samples from previous tasks. Best viewed in color.
}
\label{cilcurve}
\end{figure*}

In the paper, we reported the accuracy after all tasks have been learned. Here we give the \textit{average incremental accuracy}. Let $A_{k}$ be the average accuracy over all tasks seen so far right after the task $k$ is learned. The average incremental accuracy is defined as $\mathcal{A} = \sum_{k=1}^{t} A_{k} / t$, where $t$ is the last task. It measures the performance of a method throughout the learning process. Tab. \ref{avginc} shows the average incremental accuracy for the TIL and CIL settings. Figures \ref{tilcurve} and \ref{cilcurve} plot the TIL and CIL accuracy $A_k$ at each task $k$ for every dataset, respectively. {We can clearly see that our proposed method CLOM and CLOM(-c) outperform all others except for MNIST-5T, for which a few systems have the same results.} 

\begin{table*}[!ht]
\centering
\resizebox{2.1\columnwidth}{!}{
\begin{tabular}{l c c c c c c c c c c c c}
\toprule
\multirow{2}{*}{Method}  &  \multicolumn{2}{c}{MNIST-5T} & \multicolumn{2}{c}{CIFAR10-5T}  &  \multicolumn{2}{c}{CIFAR100-10T} &  \multicolumn{2}{c}{CIFAR100-20T} &  \multicolumn{2}{c}{T-ImageNet-5T} & \multicolumn{2}{c}{T-ImageNet-10T}\\
{} & TIL & CIL  &  TIL & CIL  &  TIL & CIL  &  TIL & CIL  &  TIL & CIL & TIL & CIL\\
\midrule
\multicolumn{13}{c}{\textbf{CIL Systems}} \\
\hline
OWM & \textbf{99.9} & \textbf{98.5} & 87.5 & 67.9 & 62.9 & 41.9 & 66.8 & 37.3 & 26.4 & 18.7 & 31.3 & 17.6 \Tstrut \\
MUC & \textbf{99.9} & 87.2 & 95.2 & 67.7 & 80.3 & 50.5 & 77.8 & 32.7 & 61.1 & 48.1 & 56.5 & 34.8\\ 
PASS  & \textbf{99.9} & 92.0 & 88.0 & 63.6 & 77.3 & 52.9 & 78.4 & 38.0 & 55.1 & 39.9 & 52.2 & 30.1\\ 
LwF.R & \textbf{99.9} & 92.8 & 96.6 & 70.7 & 87.6 & 65.4 & 90.9 & 62.8 & 61.7 & 48.0 & 61.3 & 40.9 \\
iCaRL & \textbf{99.9} & 98.0 & 96.4 & 74.7 & 86.9 & 68.4 & 88.9 & 64.5 & 60.9 & 50.7 & 60.0 & 44.1 \\
Mnemonics & \textbf{99.9} & 98.3 & 96.4 & 75.2 & 86.4 & 67.7 & 88.9 & 64.5 & 61.0 & 50.7 & 60.4 & 44.5 \\ 
BiC & \textbf{99.9} & 95.3 & 93.9 & 74.9 & 88.9 & 68.7 & 91.5 & 61.9 & 52.7 & 36.7 & 57.4 & 35.5 \\
DER++ & \textbf{99.9} & 98.3 & 94.4 & 79.3 & 86.0 & 67.6 & 85.7 & 56.6 & 62.6 & 49.7 & 66.2 & 46.8 \\
\hline
\multicolumn{13}{c}{\textbf{TIL Systems}} \Tstrut\Bstrut \\ 
\hline
HAT & \textbf{99.9} & 90.8 & 96.7 & 73.0 & 84.3 & 55.6 & 85.5 & 41.8 & 61.4 & 48.0 & 63.1 & 40.2 \Tstrut \\
HyperNet & 99.8 & 71.5 & 95.0 & 63.5 & 77.0 & 44.4 & 82.1 & 33.8 & 23.5 & 13.8 & 28.8 & 12.2 \\
SupSup & 99.7 & 45.3 & 95.5 & 50.2 & 86.1 & 49.5 & 89.1 & 28.8 & 60.6 & 45.3 & 63.4 & 37.3 \\
\hline
CLOM(-c) & \textbf{99.9} & 97.0 & \textbf{98.7} & \textbf{91.9} & \textbf{92.3} & 75.4 & \textbf{94.3} & 70.1 & \textbf{68.5} & 57.0 & \textbf{72.0} & 56.1 \Tstrut \\ 
CLOM & \textbf{99.9} & 98.3 & \textbf{98.7} & \textbf{91.9} & \textbf{92.3} & \textbf{75.9} & \textbf{94.3} & \textbf{71.0} & \textbf{68.5} & \textbf{58.6} & \textbf{72.0} & \textbf{56.5} \\
%
\bottomrule
\end{tabular}
}
\caption{Average incremental accuracy. Numbers in bold are the best results in each column.}
\label{avginc}
\end{table*}

\section{Network Parameter Sizes}\label{param_size}
We use AlexNet-like architecture \cite{NIPS2012_c399862d} for MNIST and ResNet-18 \cite{he2016deep} for CIFAR10. For CIFAR100 and Tiny-ImageNet, we use the same ResNet-18 structure used for CIFAR10, but we double the number of channels of each convolution in order to learn more tasks.

We use the same backbone architecture for CLOM and baselines, except for OWM and HyperNet, where we use the same architecture as in their original papers. OWM uses an Alexnet-like structure for all datasets. {OWN has difficulty to work with ResNet-18 because {it is not obvious how to deal with batch normalization in OWM.}} HyperNet uses a fully-connected network for MNIST and ResNet-32 for other datasets. {We found it very hard to change HyperNet because {the network initialization requires some arguments which were not explained in the paper.}} In Tab. \ref{netsize}, we report the network parameter sizes after the final task in each experiment has been trained.

Due to hard attention embeddings and task specific heads, 
CLOM requires task specific parameters for each task. For MNIST, CIFAR10, CIFAR100-10T, CIFAR100-20T, Tiny-ImageNet-5T, and Tiny-ImageNet-10T, we add task specific parameters of size 7.7K, 17.6K, 68.0K, 47.5K, 191.0K, and 109.0K, respectively, after each task. The contrastive learning also introduces task specific parameters from the projection function $g$. However, this can be discarded during deployment as it is not necessary for inference or testing.

\begin{table*}[!ht]
\centering
\resizebox{2.1\columnwidth}{!}{
\begin{tabular}{l c c c c c c}
\toprule
\multicolumn{1}{l}{Method} & \multicolumn{1}{c}{MNIST-5T} & \multicolumn{1}{c}{CIFAR10-5T} & \multicolumn{1}{c}{CIFAR100-10T} & \multicolumn{1}{c}{CIFAR100-20T} & \multicolumn{1}{c}{T-ImageNet-5T} & \multicolumn{1}{c}{T-ImageNet-10T}\\
\midrule
OWM & 5.27 & 5.27 & 5.36 & 5.36 & 5.46 & 5.46 \\
MUC-LwF & 1.06 & 11.19 & 45.06 & 45.06 & 45.47 & 45.47 \\ 
PASS & 1.03 & 11.17 & 44.76 & 44.76 & 44.86 & 44.86 \\ 
LwF.R & 1.03 & 11.17 & 44.76 & 44.76 & 44.86 & 44.86 \\
iCaRL & 1.03 & 11.17 & 44.76 & 44.76 & 44.86 & 44.86 \\
Mnemonics & 1.03 & 11.17 & 44.76 & 44.76 & 44.86 & 44.86 \\
BiC & 1.03 & 11.17 & 44.76 & 44.76 & 44.86 & 44.86 \\
DER++ & 1.03 & 11.17 & 44.76 & 44.76 & 44.86 & 44.86 \\
HAT & 1.04 & 11.23 & 45.01 & 45.28 & 44.97 & 45.11 \\
HyperNet & 0.48 & 0.47 & 0.47 & 0.47 & 0.48 & 0.48 \\
SupSup & 0.58 & 11.16 & 44.64 & 44.64 & 44.67 & 44.65  \\
CLOM & 1.07 & 11.25 & 45.31 & 45.58 & 45.59 & 45.72  \\

\bottomrule
\end{tabular}
}
\caption{Number of network parameters (million) after the final task has been learned.}
\label{netsize}
\end{table*}

\section{Details about Augmentations}\label{aug_details}
We follow \cite{chen2020simple,tack2020csi} for the choice of data augmentations. We first apply \textit{horizontal flip}, \textit{color change} (\textit{color jitter} and \textit{grayscale}), and \textit{Inception crop} \cite{inception}, and then four \textit{rotations} ($0^\circ$, $90^\circ$, $180^\circ$, and $270^\circ$). The details about each augmentation are the following. 

\begin{figure}
\centering     
\subfigure[Original]{\includegraphics[width=22.5mm]{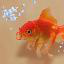}}
\subfigure[Hflip]{\includegraphics[width=22.5mm]{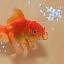}}
\subfigure[Color jitter]{\includegraphics[width=22.5mm]{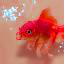}}
\subfigure[Grayscale]{\includegraphics[width=22.5mm]{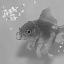}}
\subfigure[Crop]{\includegraphics[width=22.5mm]{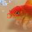}}
\subfigure[Rotation ($90^{\circ}$)
]
{\includegraphics[width=22.5mm]{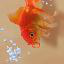}}
\caption{An original image and its view after each augmentation. Hflip and Crop refer to horizontal flip and Inception crop, respectively.}
\label{aug}
\end{figure}

\textbf{Horizontal flip}: we flip an image horizontally with 50\% of probability; \textbf{color jitter}: we add a noise to an image to change the brightness, contrast, and saturation of the image with 80\% of probability; \textbf{grayscale}: we change an image to grayscale with 20\% of probability; \textbf{Inception crop}: we uniformly choose a resize factor from 0.08 to 1.0 for each image, and crop an area of the image and resize it to the original image size; \textbf{rotation}: we rotate an image by $0^\circ$, $90^\circ$, $180^\circ$, and $270^\circ$. In Fig. \ref{aug}, we give an example of each augmentation by using an image from Tiny-ImageNet \cite{Le2015TinyIV}.

\section{Hyper-parameters} \label{hyperparam}
Here we report the hyper-parameters that we could not include in the main paper due to space limitations. 
{We use the values chosen by \cite{chen2020simple,tack2020csi} to save time for hyper-parameter search.}
We first train the feature extractor $h$ and projection function $g$ for $700$ epochs, fine-tune the classifier $f$ for $100$ epochs.
The $s$ for the pseudo step function in Eq. 7 of the main paper is set to 700. The temperature $\tau$ in contrastive loss is $0.07$ and the resize factor for Inception crop ranges from 0.08 to 1.0.

For other hyper-parameters in CLOM, {we use 10\% of training data as the validation data and select the set of hyper-parameters that gives the highest CIL accuracy on the validation set.} 
{We train the output calibration parameters $(\boldsymbol{\sigma}, \boldsymbol{\mu})$ for $160$ iterations with learning rate 0.01 and batch size 32.}
The following are experiment specific hyper-parameters found with hyper-parameter search.
\begin{itemize}
    \item For MNIST-5T, batch size = 256, the hard attention regularization hyper-parameters are $\lambda_{1} = 0.25$, and $\lambda_2 = \cdots = \lambda_{5} = 0.1$. 
    \item For CIFAR10-5T, batch size = 128, the hard attention regularization hyper-parameters are $\lambda_1 = 1.0$, and $\lambda_{2} = \cdots = \lambda_{5} = 0.75$.
    \item For CIFAR100-10T, batch size = 128, the hard attention regularization hyper-parameters are $\lambda_1 = 1.5$, and $\lambda_{2} =\cdots = \lambda_{10} = 1.0$.
    \item For CIFAR100-20T, batch size = 128, the hard attention regularization hyper-parameters are $\lambda_1 = 3.5$, and $\lambda_{2} =\cdots = \lambda_{20} = 2.5$.
    \item For Tiny-ImageNet-5T, batch size = 128, the hard attention regularization hyper-parameters are $\lambda_1 =\cdots = \lambda_{5} = 0.75$.
    \item For Tiny-ImageNet-10T, batch size = 128, the hard attention regularization hyper-parameters are $\lambda_1 = 1.0$, and $\lambda_{2} =\cdots = \lambda_{10} = 0.75$.
\end{itemize}
{We do not search hyper-parameter $\lambda_t$ for each task $t \geq 2$. However, we found that larger $\lambda_1$ than $\lambda_t$, $t>1$, results in better accuracy. This is because the hard attention regularizer $\mathcal{L}_{r}$ gives lower penalty in the earlier tasks than later tasks by definition. We encourage greater sparsity in task 1 by larger $\lambda_1$ for similar penalty values across tasks.}

{For the baselines, we use the best hyper-parameters reported in their original papers or in their code. If some hyper-parameters are unknown, e.g., the baseline did not use a particular dataset, we search for the hyper-parameters as we do for CLOM.}

We obtain the results by running the following codes

\begin{itemize}
    \item OWM: https://github.com/beijixiong3510/OWM
    \item MUC: https://github.com/liuyudut/MUC
    \item PASS: https://github.com/Impression2805/CVPR21\_PASS
    \item LwF.R: https://github.com/yaoyao-liu/class-incremental-learning
    \item iCaRL: https://github.com/yaoyao-liu/class-incremental-learning
    \item Mnemonics: https://github.com/yaoyao-liu/class-incremental-learning
    \item BiC: https://github.com/sairin1202/BIC
    \item DER++: https://github.com/aimagelab/mammoth
    \item HAT: https://github.com/joansj/hat
    \item HyperNet: https://github.com/chrhenning/hypercl
    \item SupSup: https://github.com/RAIVNLab/supsup
\end{itemize}

\end{document}